\documentclass{article}

\usepackage[final,sglblindworkshop]{neurips_2026}

\usepackage[utf8]{inputenc}
\usepackage[T1]{fontenc}
\usepackage{hyperref}
\usepackage{url}
\usepackage{booktabs}
\usepackage{amsfonts}
\usepackage{amsmath}
\usepackage{amssymb}
\usepackage{nicefrac}
\usepackage{microtype}
\usepackage{graphicx}
\usepackage{float}
\usepackage{framed}
\usepackage{subcaption}
\hypersetup{
  hidelinks,
  hypertexnames=false,
  pdftitle={EconSkills: Studying Skill Transfer and Retrieval for Web Agents on Live Economic Data},
  pdfauthor={Yinzhu Quan and Zefang Liu}
}

\setcitestyle{round}

\newcommand{\nSource}{50}
\newcommand{\nTasks}{360}
\newcommand{\nVariant}{100}
\newcommand{\nSites}{37}
\newcommand{\nRuns}{900}
\newcommand{\nSeeds}{3}
\newcommand{\maxSteps}{30}
\newcommand{\srCa}{41.0}\newcommand{\sdCa}{2.0}
\newcommand{\srCb}{7.0}\newcommand{\sdCb}{1.0}
\newcommand{\srCc}{49.3}\newcommand{\sdCc}{3.1}
\newcommand{\dSkill}{+8.3}
\newcommand{\nHelp}{27}
\newcommand{\nHurt}{12}

\title{EconSkills: Studying Skill Transfer and Retrieval for Web Agents on Live Economic Data}
\workshoptitle{Evaluation of Interactive Agents (IAEval)}
\author{%
  Yinzhu Quan \\
  Georgia Institute of Technology \\
  Atlanta, GA, USA \\
  \texttt{yquan9@gatech.edu} \\
  \And
  Zefang Liu \\
  Capital One \\
  San Jose, CA, USA \\
  \texttt{zefang.liu@capitalone.com} \\
}

\begin{document}

\maketitle

\begin{abstract}
Web agents often revisit the same sites, yet most evaluations discard the
procedures learned in earlier successful interactions. We introduce EconSkills,
a skill library and evaluation framework that distills verified EconWebArena
trajectories into parameterized standard operating procedures for retrieving live
economic data. Each skill records its scope, navigation procedure, site-specific
guidance, verification checks, and recovery steps while replacing source-instance
values with placeholders. EconSkills separates two questions: whether a known
relevant procedure transfers to a held-out task, and whether an agent can retain
that benefit when selecting from a library. In controlled transfer, matched
skills improve success over no-skill prompting and require fewer steps on paired
successes. Under the evaluated prompt formats, the parameterized skill prompt
substantially outperforms the corresponding raw-trajectory prompt. With the
full 50-skill library, retrieval is competitive with the no-skill baseline
overall and performs best on tasks with a direct family match; approximate
matches on other tasks offset these gains.
Browser trajectories further identify when procedural guidance shortens
portal-specific navigation and when semantic verification remains necessary.
These results establish that reusable economic web procedures can transfer
across task instances and provide a concrete design target for match-aware
selection and context delivery.
\end{abstract}

\section{Introduction}
\label{sec:intro}

Autonomous web agents built on large language models (LLMs) are increasingly
capable of operating real websites \citep{webvoyager2024, seeact2024} by reading
rendered pages, filling forms, following links, and returning a final answer. Economic data retrieval is a demanding and
practically important instance of this problem. Analysts routinely pull a
specific figure, an inflation print, a policy rate, an exchange rate, a
country-level indicator, from an official portal such as a central bank, a
national statistics office, or an international organization. EconWebArena\footnote{\url{https://huggingface.co/datasets/EconWebArena/EconWebArena}} \citep{liu2025econwebarena}
formalizes exactly this setting, pairing natural language questions with the live,
authoritative pages that answer them. More broadly, LLM
agents are increasingly applied to economic and business tasks, from economic
sequential reasoning \citep{econlogicqa2024} to multi-agent inventory management
\citep{invagent2024}, supply-chain risk assessment \citep{mars2026}, and
e-commerce content generation \citep{crmagent2025}. Related systems use LLM
agents for macroeconomic simulation \citep{li2024econagent} and financial
research and valuation \citep{yang2024finrobot}. These applications depend on
current, correctly identified economic evidence, yet obtaining that evidence
from official websites remains a separate interactive problem.

Despite rapid progress, standard web-agent evaluations commonly treat each task
in isolation. Tasks that
recur with a familiar structure, such as two questions on the same portal that
differ in the reference year, country, or data series, are each solved from
scratch, and whatever an agent learns while solving the first is discarded
before it reaches the second. Reflection can help an agent improve from feedback
on a task \citep{reflexion2023}, but a one-shot agent normally discards that
experience before a related task. This is wasteful, because within a portal the
route to an answer is far more stable than the answer itself. The path to an
international organization indicator, the selector sequence on a central bank
statistics page, or the export control on a national statistics office is reusable
structure that a one-shot agent discards each time. Capturing this structure once,
as a library of skills the agent can draw on, would let it reuse the route instead
of working it out again for every task; what remains open is how the agent should
select from such a library once it holds many skills and only a few fit the task at hand.
The economic setting makes this selection problem unusually exacting. A portal
can expose neighboring series with similar labels but different units,
seasonal adjustments, geographic coverage, release vintages, or effective
dates. A useful economic web skill must therefore encode not only where to
click, but also which series is being requested and how its provenance and
interpretation should be checked.

\begin{figure}[!htbp]
  \centering
  \includegraphics[width=0.95\linewidth]{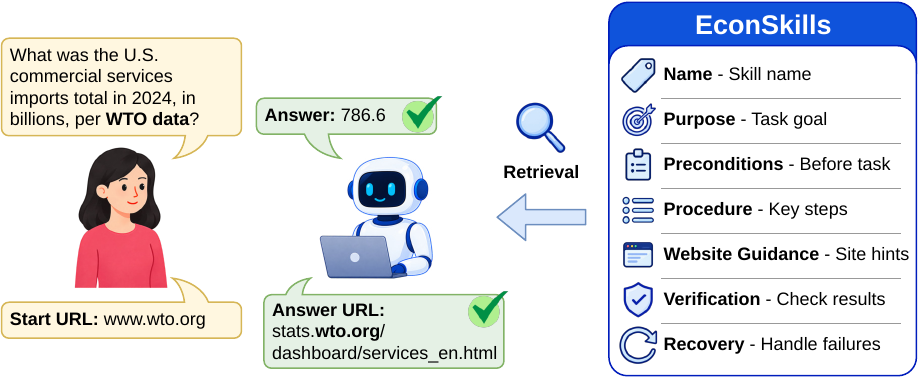}
  \caption{EconSkills overview. Reusable, instance-free skills are distilled
  from solved economic web tasks and retrieved to guide a web agent on new tasks.}
  \label{fig:pipeline}
\end{figure}

To address these gaps, we introduce EconSkills\footnote{\url{https://huggingface.co/datasets/EconWebArena/EconSkills}}, a set of abstract, instance-free
economic web skills that a web agent can draw on for new tasks, as shown in
Figure~\ref{fig:pipeline}. Building on
EconWebArena, we distill a library of \nSource{} skills, each extracted from a
solved seed trajectory as a seven-part standard operating procedure (SOP) that
records how to reach a value on an authoritative portal, covering its name,
purpose, preconditions, ordered navigation steps, site-specific guidance, and
verification and recovery rules. Instance-specific values such as a country or year
are replaced with placeholders so the procedure generalizes. We separate two
questions that are often conflated in evaluations of agent skills. First, when a
task has a known corresponding skill, does that skill transfer to a new instance?
Second, when the agent must draw from an entire library, can skill selection
preserve that benefit? We answer the first with a controlled study on
\nVariant{} held-out variants under no-skill, raw-trajectory, and matched-skill
conditions, each repeated over \nSeeds{} runs. We answer the second across all
\nTasks{} EconWebArena tasks by varying whether the agent receives no skills, all
skills, five retrieved skills, five random skills, or one optional hint.

Our contributions are threefold. (1) We introduce a two-level evaluation design
that separates procedural transfer with a known matched skill from skill
selection and delivery from a library. (2) We introduce EconSkills, a library of
\nSource{} abstract economic web skills, each a reusable SOP distilled from a
verified trajectory with instance values replaced by placeholders while retaining
preconditions, verification, and recovery. (3) We show that the two levels behave
differently: matched skills improve transfer, while full-library performance
depends on finding and presenting a suitable candidate. Paired outcomes and
trajectory analysis identify direct family match, semantic fit, and context
volume as important factors in end-to-end deployment.

\section{Related Work}
\label{sec:related}

\textbf{Web agents and benchmarks.} WebArena \citep{zhou2023webarena} evaluates
long-horizon tasks on fully interactive websites, while Mind2Web
\citep{deng2023mind2web} studies generalization across diverse real-world sites.
VisualWebArena \citep{koh2024visualwebarena} adds visually grounded tasks, and
WebLINX \citep{lu2024weblinx} focuses on conversational navigation. BrowserGym
\citep{browsergym2025} provides a common environment for running and comparing
agents across benchmarks. Evaluation has also moved toward online settings and
agentic search through WebCanvas \citep{pan2024webcanvas} and Mind2Web 2
\citep{gou2025mind2web2}. EconWebArena \citep{liu2025econwebarena} specializes
this line to multimodal, multi-step economic tasks on live, authoritative sites,
where related task families often share a procedural structure.

\textbf{Experience and workflow reuse.} Several methods carry information across
tasks rather than restarting from an empty context. ExpeL \citep{zhao2024expel}
extracts natural-language insights from prior experience and recalls relevant
examples at inference time. Agent Workflow Memory \citep{wang2025awm} induces
reusable web workflows in both offline and online settings and selectively
supplies them on new tasks, while RaDA \citep{kim2024rada} retrieves exemplars
separately for web-task decomposition and action generation. VideoWebArena
\citep{jang2024videowebarena} further shows that long-context tutorials can hurt
skill-retention performance, suggesting that procedural context is not uniformly
helpful. These results motivate cross-task reuse, but the stored procedure and
its delivery are often evaluated together. Our two-level design instead
separates the transfer value of a known match from the problem of selecting and
presenting it from a library.

\textbf{Skills for agents.} Voyager \citep{voyager2023} maintains an executable
code-skill library, and SkillWeaver \citep{zheng2025skillweaver} discovers and
refines reusable web APIs through exploration. More recent work studies
auditable and generalizable skill artifacts: SkillGen \citep{skillgen_ma2026}
verifies synthesized skills by their effect on held-out tasks, ContractSkill
\citep{contractskill2026} adds explicit verification and repair, and PolySkill
\citep{polyskill2025} targets cross-site generalization through polymorphic
abstraction. SkillsBench \citep{skillsbench2026} likewise evaluates skills
through downstream agent behavior rather than textual plausibility. SkillGen and
SkillsBench both expose task-level regressions alongside aggregate gains,
motivating paired outcome analysis rather than reporting average improvement
alone. EconSkills builds on this functional view while testing how the same skill
library behaves under both matched transfer and full-library delivery.

\section{EconSkills}
\label{sec:method}

EconSkills is built and applied in three stages, as illustrated in Figure~\ref{fig:pipeline}. Task execution
produces a solved seed trajectory, skill extraction distills that trajectory
into an abstract standard operating procedure (SOP), and skill application supplies skills from the resulting
library to a web agent and evaluates them across the EconWebArena benchmark. The library
holds \nSource{} skills, one per solved seed task, each capturing a distinct
economic data-retrieval workflow on an authoritative portal.

\subsection{Task Execution}
We solve each seed task by driving the live portal through a browser.
At every step the agent observes a structured view of the current page, its
accessibility tree (AXTree), a screenshot, and
contextual metadata such as the focused element and the action history, and it
responds with one or more actions from EconWebArena's BrowserGym\footnote{\url{https://github.com/ServiceNow/BrowserGym}} high-level
action set. This action space supports element-based and coordinate-based
interaction, keyboard input, scrolling, navigation, and tab management. The
agent repeats this loop until it commits a final answer. We keep a seed task only
after it is solved and its answer clears
the benchmark's automatic check, which requires both the correct value and a
landing URL on the expected authoritative domain. Grounding every retained skill in a
verified success, rather than in an aspirational account of how a portal ought to
behave, ensures that each skill is later distilled from a trajectory that actually
reached the target value. This procedure yields \nSource{} solved seed tasks,
one for each skill in the library. These \nSource{} seed tasks span nine of EconWebArena's ten economic categories,
namely government, banking, markets, labor, finance, energy, education, health, and
trade, and draw on \nSites{} distinct authoritative websites, including central
banks, national statistics offices, international organizations, and financial
data portals. Table~\ref{tab:corpus} lists the number of seed tasks in each category
together with the instance fields that each skill replaces with placeholders.

\begin{table}[!htbp]
  \caption{EconSkills seed tasks by category. Placeholder fields indicate the
  instance-specific values abstracted by each skill.}
  \label{tab:corpus}
  \centering
  \small
  \begin{tabular}{@{}lrp{7cm}@{}}
    \toprule
    \textbf{Category} & \textbf{Seed tasks} & \textbf{Placeholder fields} \\
    \midrule
    Government & 18 & \texttt{country}, \texttt{state}, \texttt{date}, \texttt{indicator}, \texttt{group}, \texttt{bracket} \\
    Banking    & 13 & \texttt{date}, \texttt{indicator}, \texttt{instrument} \\
    Markets    & 9  & \texttt{date}, \texttt{instrument}, \texttt{price}, \texttt{volume} \\
    Labor      & 3  & \texttt{state}, \texttt{sex}, \texttt{level}, \texttt{occupation} \\
    Finance    & 2  & \texttt{state}, \texttt{count}, \texttt{rate}, \texttt{bracket} \\
    Energy     & 2  & \texttt{date} \\
    Education  & 1  & \texttt{item} \\
    Health     & 1  & \texttt{bracket}, \texttt{filing} \\
    Trade      & 1  & \texttt{exporter}, \texttt{importer} \\
    \midrule
    Total      & \nSource{} & \nSites{} websites across nine categories \\
    \bottomrule
  \end{tabular}
\end{table}

\subsection{Skill Extraction}
From each solved trajectory, \texttt{gpt-5} distills one structured skill,
written as a seven-part SOP using a fixed instruction and one in-context example.
The extractor reads the full trajectory that solved
the seed task, the visited URLs, the AXTree observations, the actions
taken, and the confirmed final answer, and rewrites it into a procedure that a
different agent can follow on a fresh instance. Its JSON output is parsed and
validated against the seven-field schema before inclusion in the library. The
seven fields are (i) name,
(ii) purpose, the kind of questions the skill answers, (iii) preconditions,
what must hold before the procedure begins, such as the portal and dataset being
reachable, (iv) procedure, the ordered navigation and interaction steps, (v)
website guidance, the portal-specific menus, selectors, and pitfalls, (vi)
verification, how to confirm that the retrieved figure is the one the task
requested, and (vii) recovery, how to react when a step fails.

The last two fields are especially important for economic data. Official
portals frequently place nominal and real values, adjusted and unadjusted
series, preliminary and revised releases, or multiple effective dates on the
same page. EconSkills therefore records checks for entity, period, unit, series
definition, and source domain, together with portal-specific fallbacks such as
table, archive, and download views. This domain information is retained even
when the instance values themselves are abstracted.

The property that makes a skill reusable is abstraction. The extractor identifies
every instance-specific value the seed trajectory touched, a year, a country, an
indicator, or a numeric target, and replaces it with a named placeholder such as
\texttt{<year>}, \texttt{<country>}, or \texttt{<indicator>}. The procedure then
refers to these slots rather than to one concrete question, so a step such as
select the \texttt{<indicator>} series, set the region to \texttt{<country>}, and
set the period to \texttt{<date>} stays valid when the placeholders are filled for
a different instance. Verification and recovery are what separate a skill from a
brittle recording of clicks. The verification field states how to confirm that the
value read back matches the requested units, period, and entity, and the recovery
field states what to do when a step fails, for instance when a menu label has
changed or a selector is missing, by falling back to the portal's search or an
alternate export view. We serialize each skill as \texttt{JSON} over these seven
fields and render it to Markdown for injection into the web agent, and we
extract exactly one skill per solved seed task, which gives \nSource{} independent
skills. Table~\ref{tab:example-skill} gives an example skill, extracted from a
European Central Bank exchange-rate seed task, with the counter-currency and date
reduced to placeholders. Appendix~\ref{app:repro} gives library and reproduction
details, and Appendix~\ref{app:skill-example} reproduces this example
and a second skill in full.

\begin{table}[!htbp]
  \caption{Example seven-part skill extracted from an ECB exchange-rate seed
  task. Counter-currency and date are represented as placeholders.}
  \label{tab:example-skill}
  \centering
  \small
  \begin{tabular}{@{}lp{9.4cm}@{}}
    \toprule
    \textbf{Field} & \textbf{Content} \\
    \midrule
    Name             & \texttt{ecb\_reference\_fx\_rate\_single\_date} \\
    Purpose          & Retrieve the ECB euro foreign-exchange reference rate for a counter-currency on a given date. \\
    Preconditions    & The task asks for an ECB-reported rate; the base is EUR with a counter-currency and a calendar date; the answer must come from an \texttt{ecb.europa.eu} page. \\
    Procedure        & Open the ECB euro reference-rates section; open the Euro-to-\texttt{<currency>} page; set both date fields to \texttt{<date>}; apply; read the reference rate for that day. \\
    Website guidance & Currency pages follow \texttt{eurofxref-graph-<iso>.en.html}; set From and To to the same date; use the CSV download if the chart is ambiguous. \\
    Verification     & Confirm the EUR/\texttt{<currency>} orientation is not inverted, the date and currency match the request, and the value is the daily reference rate rather than a market spot. \\
    Recovery         & If the currency page is missing, pick it from the landing page or site-search \texttt{ecb.europa.eu}; if date entry is rejected, use the date picker; if no value exists on \texttt{<date>}, check whether it is a non-business day. \\
    \bottomrule
  \end{tabular}
\end{table}

\subsection{Skill Application}
\label{sec:application}
We evaluate EconSkills at two levels. The first isolates whether a
correctly matched procedure transfers; the second asks whether an agent can use
the whole library when the match is not given.

\textbf{Controlled matched-skill transfer.}
Each of the \nSource{} source skills is paired with two held-out task variants
from the same procedural family, yielding \nVariant{} source-variant pairs. The
variants change an instance field such as date, entity, series, or category while
preserving the website workflow. For each variant and condition, we conduct
\nSeeds{} independent agent runs, using seeds 0, 1, and 2: \textbf{BASE}, the task alone;
\textbf{TRAJ1}, the task plus one matched raw successful trajectory with its
final answer removed; and \textbf{MATCH1}, the task plus one directly matched
extracted skill. Thus BASE versus MATCH1 tests transfer from a matched skill,
while TRAJ1 versus MATCH1 compares the packaged skill representation with a
verbose task-specific history. Because these prompts differ in abstraction,
length, and format, this comparison does not isolate abstraction alone.

\textbf{Full-library deployment.}
We next run all \nTasks{} tasks once under conditions that vary which EconSkills
library context is supplied: \textbf{BASE} receives no skill; \textbf{ALL50}
receives the full library; \textbf{RETR5} receives the five skills ranked most
relevant using their names and purposes rather than their full procedures; and
\textbf{RAND5} receives five task-seeded random
skills. RETR5 and RAND5 hold context count fixed, while ALL50 tests whether
additional procedures help or interfere.

We additionally evaluate \textbf{HINT30}, which injects one skill but explicitly
tells the agent that it is an optional, possibly outdated reference that may be
ignored whenever it does not match the page. For the 150 tasks in the 50
source families (50 sources and 100 variants), HINT30 supplies the directly
corresponding family skill. The remaining 210 tasks have no direct family match
and receive a top-1 retrieved skill. We analyze these groups separately because
mixing them would confound transfer with candidate selection. This label is
relative to the 50 source families, not an external annotation that no other
reusable procedure could apply.

\textbf{Metrics.}
For task $t$ and condition $c$, let $r_c(t)\in\{0,1\}$ be the native
EconWebArena reward, which is one only when the answer contains the gold value
and the final URL lies on the authoritative domain. We report
$\mathrm{SR}_c=N^{-1}\sum_t r_c(t)$ and interaction steps. Matched outcomes are
compared with McNemar's exact test~\citep{mcnemar1947}; step comparisons are
restricted to runs where both conditions succeed, so early failure is not
mistaken for efficiency.

\section{Experiments}
\label{sec:experiments}

We first test transfer where the correct skill is known, then test deployment
where the agent must draw from the library. The two experiments measure different
parts of the system: the first evaluates the stored procedure, whereas the second
also depends on deciding which procedure belongs with a new task.

\subsection{Experimental Setup}
\label{sec:setup}

EconWebArena contains \nTasks{} live tasks on authoritative portals across
government, banking, markets, labor, finance, energy, education, health, and
trade. Our controlled study uses the \nVariant{} held-out variants with a direct
match to the \nSource{} skill families; our full-library deployment study uses all
\nTasks{} tasks. No task requires API credentials, and every answer is obtained
through browser interaction with publicly reachable pages.

Our agent is instantiated on
\texttt{gpt-5-mini}\footnote{\url{https://developers.openai.com/api/docs/models/gpt-5-mini}} within
the BrowserGym and
AgentLab\footnote{\url{https://github.com/ServiceNow/AgentLab}} ecosystem~\citep{browsergym2025}. At each step it receives the page accessibility tree (AXTree), a
screenshot, and contextual metadata, namely the focused element and the recent
action history, and emits one or more actions from the EconWebArena high-level
action set, terminating when it returns a final answer or reaches the
\maxSteps{}-step budget. The agent
configuration, observation space, and step budget are held fixed across all
conditions. The experimental conditions differ only in the trajectory or skill
text added to the task prompt.

We compare the experimental conditions defined in
Section~\ref{sec:application}. For each of the \nVariant{} variants, BASE,
TRAJ1, and MATCH1 are each evaluated in \nSeeds{} independent runs, using seeds
0, 1, and 2. This yields \nVariant{} variants $\times$ 3 conditions $\times$
\nSeeds{} runs = \nRuns{} browser-agent episodes. Each deployment condition
contains one run per task across all \nTasks{} tasks. Supplied trajectories or skills are serialized to Markdown and
prepended through the same extra-instructions channel. Within each study, the
model, browser environment, observations, and action budget are fixed. We use the
native benchmark reward and do not rescue answers from the agent text after a
failed automatic check. Four independently verified benchmark values that had
changed at their authoritative source are handled through a fixed correction
table constructed without reference to condition outcomes; details and
the reproduction protocol are in Appendix~\ref{app:repro}.

\subsection{Experimental Results}
\label{sec:results}

We first evaluate EconSkills under controlled transfer with a known skill match,
then under full-library deployment where the agent must select its own context.
Subsequent analyses examine paired outcomes, representative browser cases, and
direct family matches.

\subsubsection{RQ1: Do matched skills transfer to held-out tasks?}

Table~\ref{tab:controlled} reports the controlled transfer study. Supplying the
matched extracted skill raises success from \srCa\% to \srCc\%, a gain of
\dSkill{} percentage points. On the 300 matched seed-level outcomes, MATCH1 changes
46 BASE failures into successes, compared with 21 changes in the opposite direction;
McNemar's exact test gives $p = 0.0031$. This positive imbalance confirms that the
gain is distributed across paired outcomes rather than driven only by aggregate
averaging, while also motivating the case analysis in RQ3.
Appendix~\ref{app:audit} provides the cell-level aggregation and testing details.

\begin{table}[!htbp]
  \caption{Controlled matched-skill transfer on \nVariant{} held-out variants.
  Each variant-condition pair has 3 independent runs, using seeds 0, 1, and 2;
  standard deviations are across the 3 run-level success rates. The highest success values and lowest mean step
  count are bold.}
  \label{tab:controlled}
  \centering
  \small
  \begin{tabular}{@{}lcrr@{}}
    \toprule
    \textbf{Condition} & \textbf{Success} & \textbf{SR (\%)} & \textbf{Mean steps} \\
    \midrule
    BASE: No skill & 123 / 300 & $\srCa \pm \sdCa$ & 19.83 \\
    TRAJ1: One matched raw trajectory & \phantom{0}21 / 300 & $\srCb \pm \sdCb$ & \textbf{7.95} \\
    MATCH1: One matched skill & \textbf{148 / 300} & $\mathbf{\srCc \pm \sdCc}$ & 15.67 \\
    \bottomrule
  \end{tabular}
\end{table}

By contrast, the raw source trajectory reaches only \srCb\%, although it comes
from a successful task in the same family. Its short mean episode length is
caused by early termination rather than efficient completion. Simply deleting
the source answer, while leaving instance-specific actions and observations in
place, does not produce a useful demonstration for the new instance. The result
favors the structured, parameterized skill prompt over this raw-trajectory
prompt under the evaluated formats. It does not attribute the entire difference
to abstraction because prompt length and presentation also change.

We compare step counts only for the 102 seed-level cells in which both BASE and MATCH1
succeed, since a failed run often uses the entire \maxSteps{}-step budget. Within
this subset, MATCH1 uses 2.37 fewer actions on average (median two fewer). It is
shorter in 59 cells, tied in 15, and longer in 28 (two-sided sign test excluding
ties, $p = 0.0012$).

\subsubsection{RQ2: How does skill delivery change when selecting from a library?}

Table~\ref{tab:library} reports the experiment in which the correct skill is no
longer given. RETR5 is the strongest full-library condition and is statistically
tied with BASE overall (28.3\% and 28.1\%; McNemar $p = 1.0$). On direct-match
tasks, it also leads all conditions at 48.0\%, compared with 44.0\% for BASE.
RAND5 reaches 24.4\% ($p = 0.085$ against BASE), while ALL50 reaches 11.4\%
($p < 10^{-12}$). Together, these results show that a compact retrieved set
preserves baseline-level overall performance and leads on the direct-match subset,
whereas indiscriminate full-library delivery introduces substantial context
interference.

\begin{table}[!htbp]
  \caption{EconSkills full-library deployment by direct family match. Cells
  report successes / tasks (SR). Each condition has one run per task; the best
  result in each column is bold and the second-best is underlined.}
  \label{tab:library}
  \centering
  \small
  \begin{tabular}{@{}lrrr@{}}
    \toprule
    \textbf{Condition} & \textbf{Direct match (150)} & \textbf{No direct match (210)} & \textbf{Overall (360)} \\
    \midrule
    BASE: No skill & 66 / 150 (44.0\%) & \textbf{35 / 210 (16.7\%)} & \underline{101 / 360 (28.1\%)} \\
    ALL50: Full library & 27 / 150 (18.0\%) & 14 / 210 (6.7\%) & 41 / 360 (11.4\%) \\
    RETR5: Five retrieved & \textbf{72 / 150 (48.0\%)} & \underline{30 / 210 (14.3\%)} & \textbf{102 / 360 (28.3\%)} \\
    RAND5: Five random & 62 / 150 (41.3\%) & 26 / 210 (12.4\%) & 88 / 360 (24.4\%) \\
    HINT30: One optional hint & \underline{69 / 150 (46.0\%)} & 24 / 210 (11.4\%) & 93 / 360 (25.8\%) \\
    \bottomrule
  \end{tabular}
\end{table}

The direct-match split clarifies the aggregate comparison: RETR5 and HINT30 gain
on the 150 tasks drawn from a source family, while all skill conditions trail BASE
on the 210 tasks without such a match. Thus RETR5's overall tie combines gains
from known families with losses when candidate relevance is uncertain.

HINT30 asks whether explicitly allowing the agent to ignore one optional skill
reduces this interference. Its direct-match difference from BASE is not
significant ($p = 0.711$), and both conditions solve 45 of the 100 held-out
variants. On tasks without a direct family match, the approximate top-1 hint lowers success from
16.7\% to 11.4\% ($p = 0.043$). Across all 360 tasks, HINT30 obtains 25.8\%,
compared with 28.1\% for BASE ($p = 0.341$).

We next move from aggregate rates to paired outcomes and preserved trajectories,
asking which task and interface properties govern successful reuse.

\subsubsection{RQ3: What determines successful skill reuse?}
\label{sec:analysis}

\begin{table}[!htbp]
  \caption{Pair-level outcomes in the controlled study. Categories compare mean
  success across the 3 runs for each source-variant pair.}
  \label{tab:pair-outcomes}
  \centering
  \small
  \begin{tabular}{@{}lrr@{}}
    \toprule
    \textbf{Outcome pattern} & \textbf{Pairs} & \textbf{Percentage} \\
    \midrule
    Matched skill improves over no skill & 27 & 27\% \\
    Matched skill reduces success & 12 & 12\% \\
    Unchanged, with at least one success & 29 & 29\% \\
    No success in any condition or run & 32 & 32\% \\
    \midrule
    All & 100 & 100\% \\
    \bottomrule
  \end{tabular}
\end{table}

Table~\ref{tab:pair-outcomes} summarizes the pair-level outcomes. The matched
skill improves the mean outcome for \nHelp{} of the \nVariant{} source-variant
pairs, more than twice the \nHurt{} pairs that move in the opposite direction;
29 pairs have equal mean success with at least one successful run. The improvements
tend to occur when a portal has a stable
series page or selector sequence. Examples include ONS inflation time series,
Treasury yield tables, and CoinMarketCap historical snapshots. A match at the
website or task-family level is nevertheless not always sufficient. The boundary
cases involve a nearby but different semantic choice, such as a form-specific
fee category or the distinction between close and adjusted close.

We also read the preserved trajectories from the full-library deployment and
HINT30 conditions. They identify
two boundaries on reuse. First, a dynamic selector, download control, or
redesigned page can prevent the agent from following a recorded route. Second,
an agent may reach the relevant page but read the wrong unit, date, series, or
filing category. We further observe runs in which an approximately relevant
skill keeps the agent on the wrong page after progress has stalled. These cases
help interpret HINT30 on the 210 tasks without a direct family match and motivate
explicit match checks before a candidate enters the prompt.

\begin{figure}[!htbp]
  \centering
  \setlength\fboxsep{1pt}
  \setlength\fboxrule{0.5pt}
  \begin{subfigure}[t]{0.32\linewidth}
    \centering\fbox{\includegraphics[width=.98\linewidth]{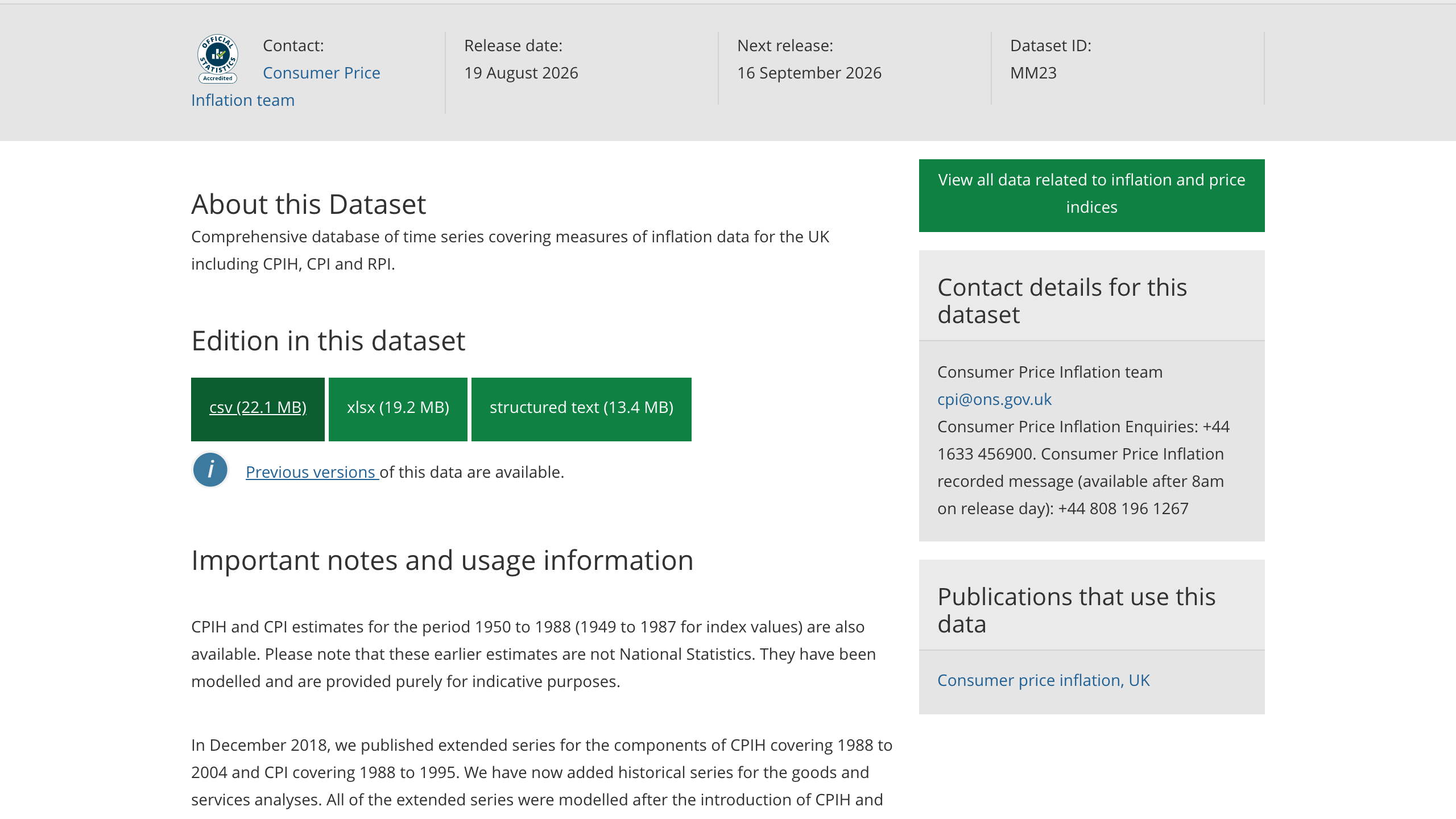}}
    \caption{ONS BASE: dataset page}
  \end{subfigure}\hfill
  \begin{subfigure}[t]{0.32\linewidth}
    \centering\fbox{\includegraphics[width=.98\linewidth]{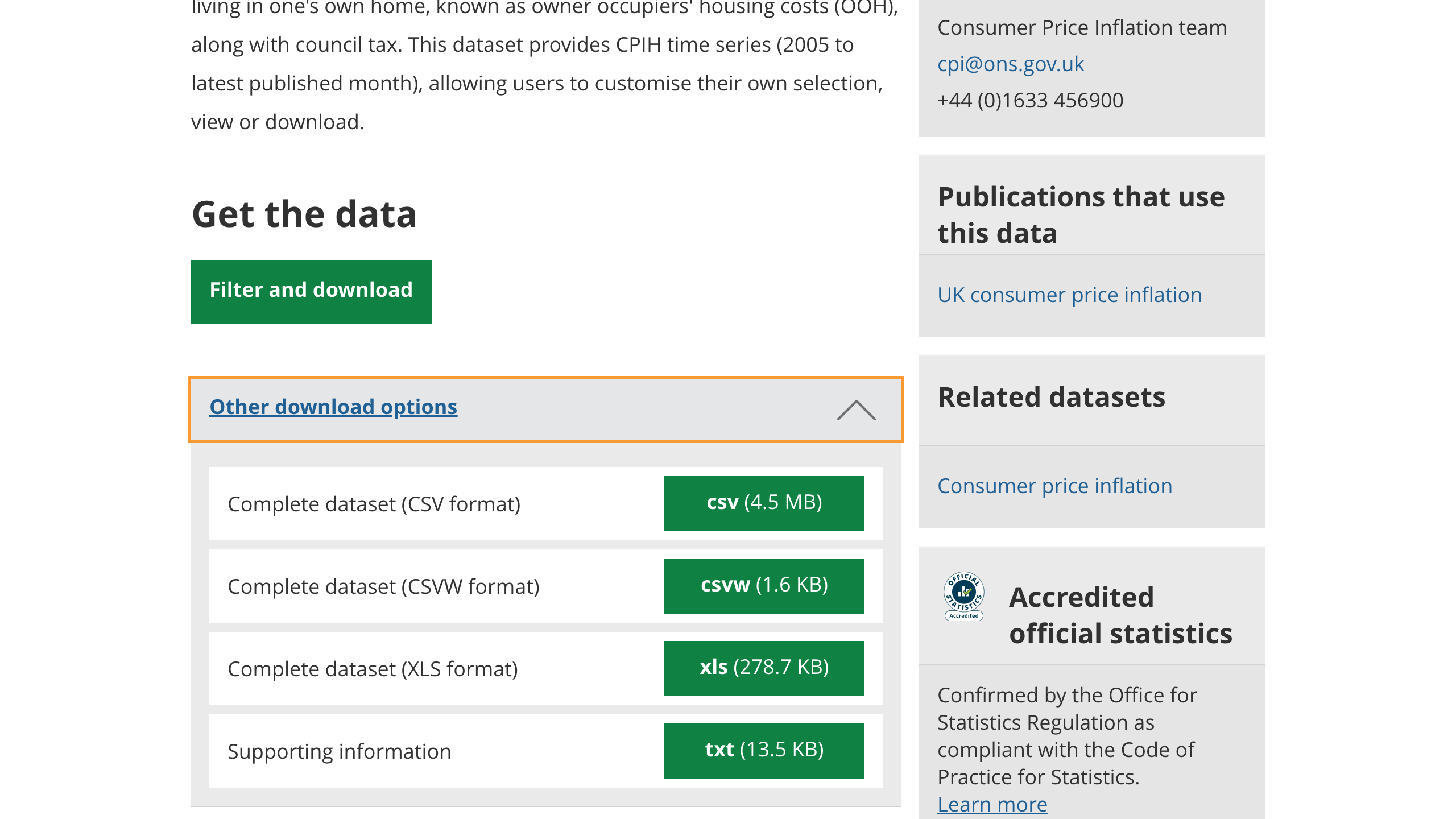}}
    \caption{ONS BASE: download view}
  \end{subfigure}\hfill
  \begin{subfigure}[t]{0.32\linewidth}
    \centering\fbox{\includegraphics[width=.98\linewidth]{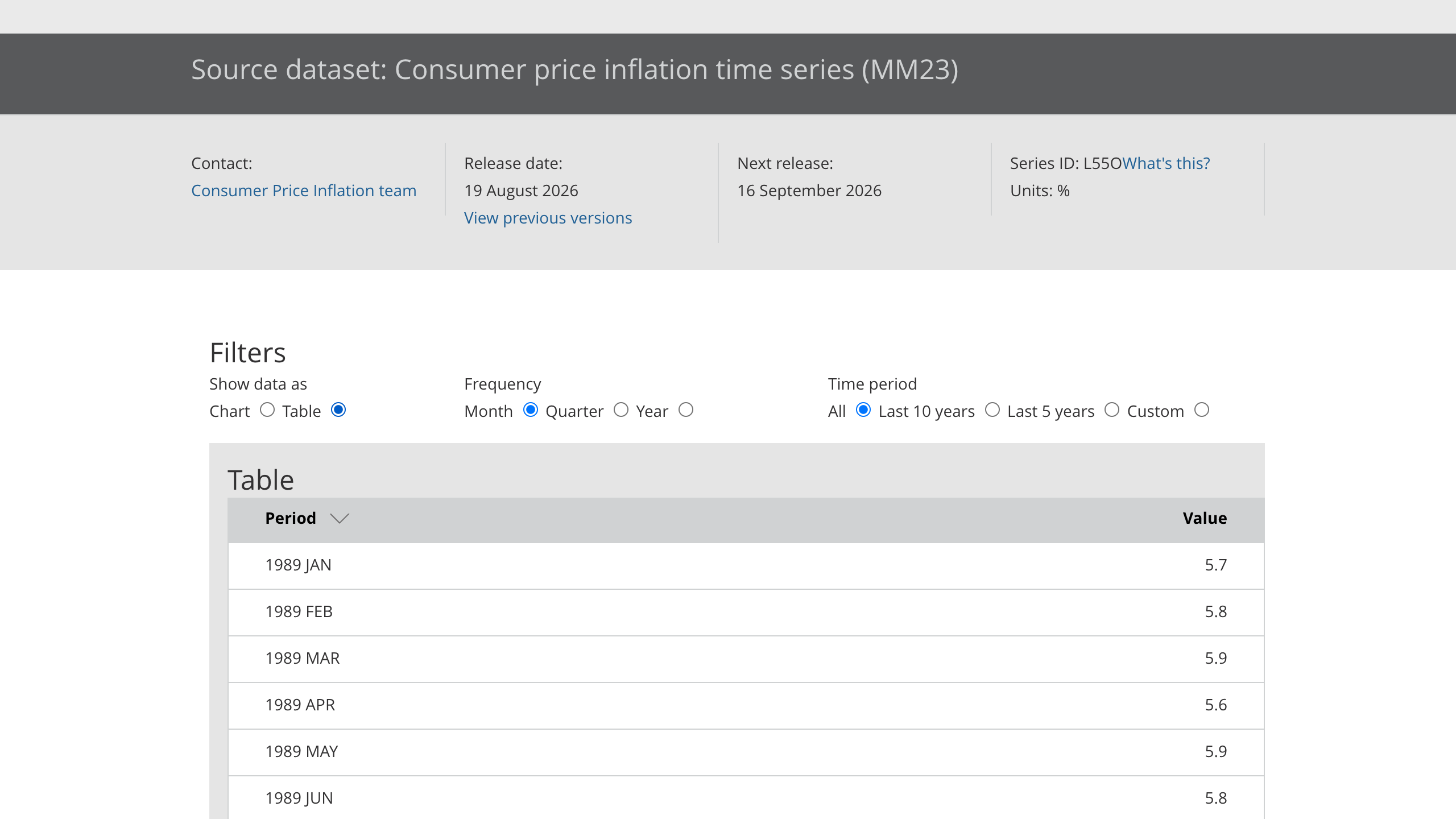}}
    \caption{ONS HINT30: answer table}
  \end{subfigure}

  \vspace{3pt}
  \begin{subfigure}[t]{0.32\linewidth}
    \centering\fbox{\includegraphics[width=.98\linewidth]{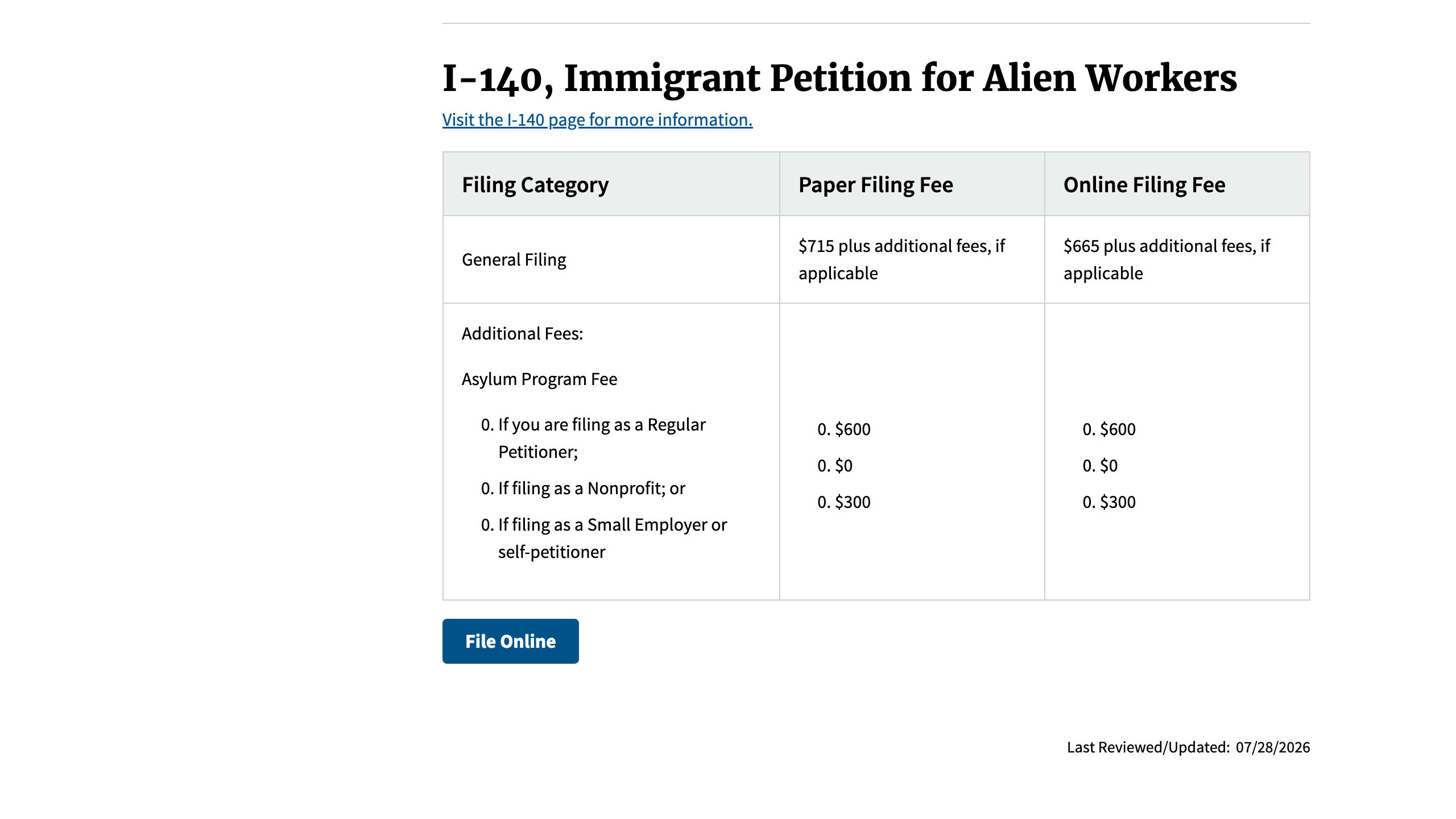}}
    \caption{USCIS BASE: answer table}
  \end{subfigure}\hfill
  \begin{subfigure}[t]{0.32\linewidth}
    \centering\fbox{\includegraphics[width=.98\linewidth]{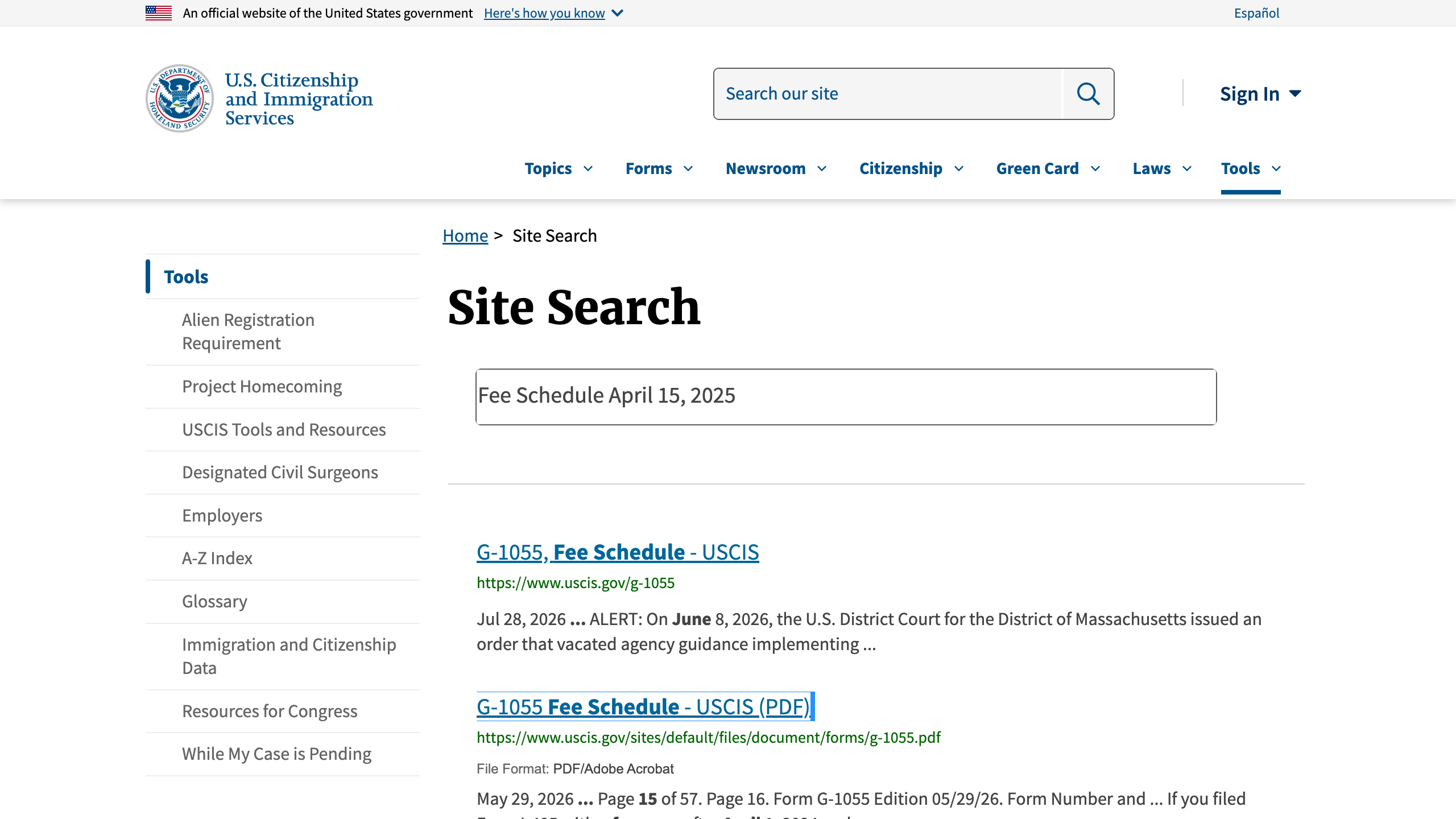}}
    \caption{USCIS HINT30: fee search}
  \end{subfigure}\hfill
  \begin{subfigure}[t]{0.32\linewidth}
    \centering\fbox{\includegraphics[width=.98\linewidth]{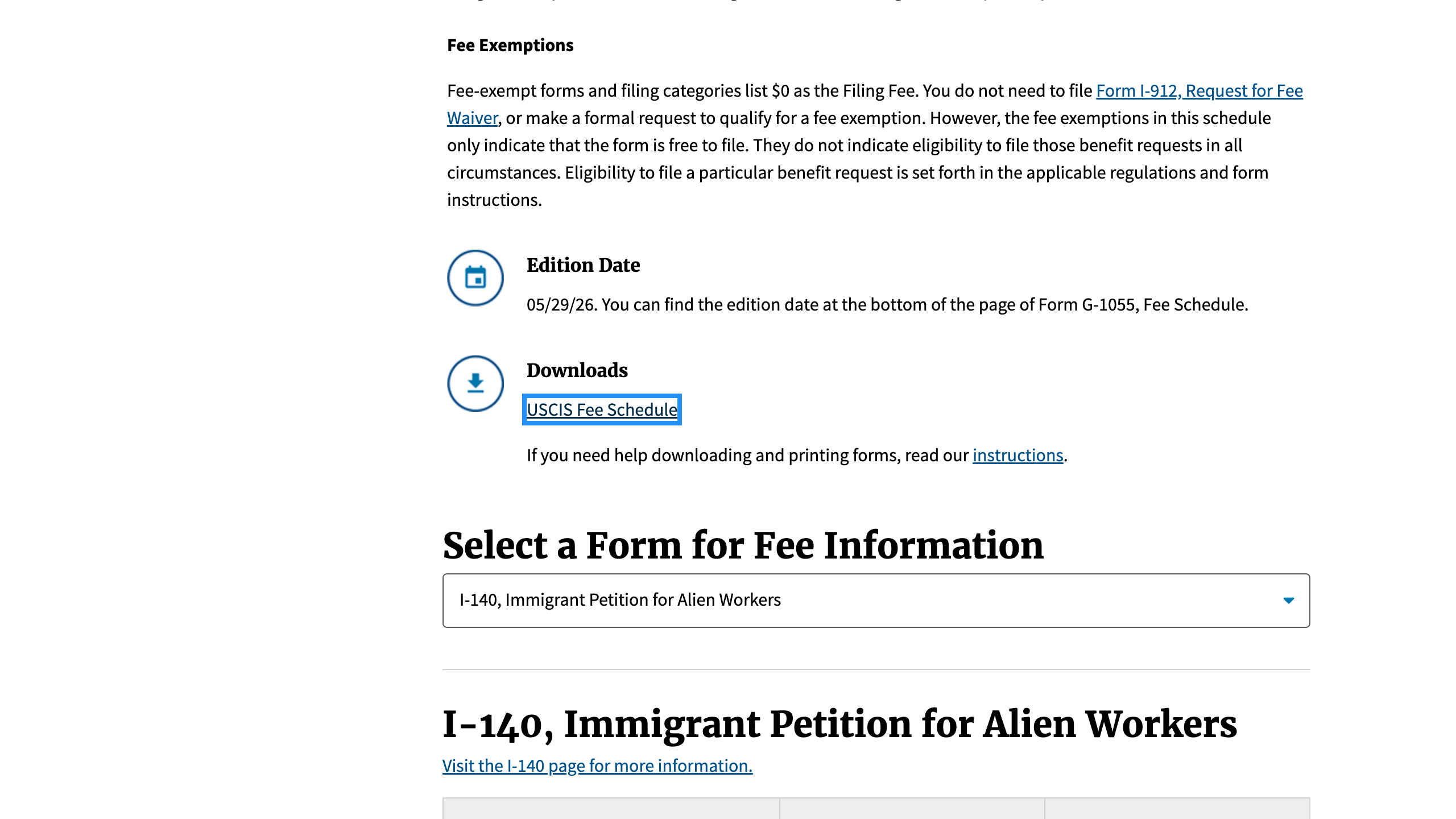}}
    \caption{USCIS HINT30: I-140 section}
  \end{subfigure}
  \caption{Representative browser states from the full-library deployment study. Top: the matched
  hint moves the ONS agent from broad dataset and download views to the dedicated
  series table. Bottom: both USCIS agents reach relevant content, but the hinted
  run does not return the verified value.}
  \label{fig:body-cases}
\end{figure}

Figure~\ref{fig:body-cases} gives a successful transfer and a boundary case. On ONS
task 225, BASE spends 30 steps at a general download interface, while HINT30
reaches the dedicated MM23 time-series page and succeeds in 14 steps. On USCIS
task 264, by contrast, the hint reaches the relevant I-140 fee section but the
agent does not return a verified answer within 30 steps. The second case shows that
reaching the right page and extracting the requested value are separate parts of
the task. Full step sequences appear in Figures~\ref{fig:ons-case}
and~\ref{fig:uscis-case} in Appendix~\ref{app:cases}.

\subsection{Direct-Match Analysis}
\label{sec:coverage}

The HINT30 aggregate combines three substantively different groups.
Table~\ref{tab:coverage} separates the source tasks used to create the skills, their
held-out variants, and tasks for which the library has no direct family match. Source
tasks are included as a diagnostic of prompt use, not as transfer evidence. The
held-out variants are the relevant transfer group, and HINT30 does not change
their aggregate success: both conditions solve 45 of 100 tasks. This differs
from the controlled MATCH1 result because HINT30 changes both the presentation and
the run protocol: the skill is framed as optional and each task is run once.

\begin{table}[!htbp]
  \caption{HINT30 and BASE outcomes by direct family match. Source tasks diagnose reuse
  on the originating tasks; held-out variants provide the transfer comparison.
  The higher success result within each subset is bold; ties are both bold.}
  \label{tab:coverage}
  \centering
  \small
  \begin{tabular}{@{}lrrrrr@{}}
    \toprule
    \textbf{Subset} & \textbf{Tasks} & \textbf{BASE} & \textbf{HINT30} & \textbf{$\Delta$ pp} & \textbf{$p$} \\
    \midrule
    Source tasks & 50 & 21 (42.0\%) & \textbf{24 (48.0\%)} & +6.0 & 0.581 \\
    Held-out variants & 100 & \textbf{45 (45.0\%)} & \textbf{45 (45.0\%)} & 0.0 & 1.000 \\
    Direct family match, total & 150 & 66 (44.0\%) & \textbf{69 (46.0\%)} & +2.0 & 0.711 \\
    No direct match & 210 & \textbf{35 (16.7\%)} & 24 (11.4\%) & -5.3 & 0.043 \\
    \bottomrule
  \end{tabular}
\end{table}

This split is defined by benchmark construction: 150 tasks belong to one of the
50 source families, while the other 210 do not. It is therefore a library-relative
label, not an external inventory of every procedure a task could use. The latter
group tests whether retrieval and the optional-hint instruction can reject an
approximate candidate when no same-family skill exists. Together with the
controlled study, the result supports a precise conclusion: the library contains
transferable procedures, while end-to-end gains depend on filtering candidates
before prompt construction.

\subsection{Discussion}
\label{sec:discussion}

The contrast between TRAJ1, MATCH1, and ALL50 shows why the form and amount of
experience matter. A raw trajectory mixes reusable steps with instance-specific
state. A full library instead makes many plausible but inapplicable procedures
compete for attention. This pattern is consistent with context interference,
although we do not isolate prompt length, conflicting instructions, and relevance
judgments. Effective reuse therefore requires explicit skill scope and selective
prompt construction.

The direct-match split suggests a concrete design target. RETR5 leads BASE when
a same-family skill exists; the overall tie includes tasks without such a match.
An operational system should therefore withhold skills when no suitable match is
found by checking the portal, data type, and interaction schema against skill
preconditions. Verification should then confirm the final page, unit, date, entity,
series definition, adjustment status, release vintage, and quote orientation when
relevant.

The current collection is large enough to expose interference from irrelevant
context, but 50 skills do not establish scalability to organizational libraries
with hundreds or thousands of procedures. Such systems will likely require
hierarchical retrieval, explicit rejection, and bounded prompt construction.
They also require maintenance: abstracting dates and values prevents skills from
caching stale answers, but portal routes and series definitions can still change.
Last-verified metadata, failure-triggered revalidation, and revision or retirement
of stale procedures are natural extensions; we do not evaluate that lifecycle.

\section{Conclusion}
\label{sec:conclusion}

EconSkills demonstrates that verified web interactions can be distilled into
reusable procedural knowledge for live economic data. In controlled transfer,
the parameterized skill prompt outperforms no-skill prompting and the evaluated
raw-trajectory prompt on held-out variants and shortens paired successful
interactions. With the full 50-skill library, RETR5 matches BASE overall and
leads it numerically when a direct family match exists. Approximate candidates
on other tasks offset these gains, while full-library delivery shows the cost of
excessive context. EconSkills thus makes skill reuse measurable as procedural
transfer and candidate selection, with concrete
checks for portal, series, date, unit, entity, and provenance. This framework
points toward web agents that accumulate and reliably reuse economic-domain
experience.

\section*{Limitations}
\label{sec:limitations}

We study one benchmark, one agent backbone, and a fixed 30-step budget. The
controlled matched-skill study assumes that the corresponding family skill is
known, so it isolates procedural transfer rather than retrieval; the
full-library deployment study instead combines library composition, candidate selection,
and skill use.
Each full-library deployment condition has one run per task, enabling paired comparisons
but not across-run variance estimates. The 50 skills come from verified
trajectories under one extraction schema, so results may change with other
models, larger libraries, or evolving sites. TRAJ1 and MATCH1 also differ in
prompt length and format, so their comparison evaluates the two packaged
representations rather than abstraction alone. Direct family match is defined
against this library and is not an external coverage annotation. Future work
should repeat deployment across seeds and test skill rejection and maintenance
under larger libraries and site change.

\small
\bibliographystyle{plainnat}
\bibliography{refs}

@inproceedings{zhou2023webarena,
  title     = {{WebArena: A Realistic Web Environment for Building Autonomous Agents}},
  author    = {Zhou, Shuyan and Xu, Frank F. and Zhu, Hao and Zhou, Xuhui and Lo, Robert and Sridhar, Abishek and Cheng, Xianyi and Ou, Tianyue and Bisk, Yonatan and Fried, Daniel and Alon, Uri and Neubig, Graham},
  booktitle = {International Conference on Learning Representations (ICLR)},
  year      = {2024},
  note      = {arXiv:2307.13854}
}

@inproceedings{deng2023mind2web,
  title     = {{Mind2Web: Towards a Generalist Agent for the Web}},
  author    = {Deng, Xiang and Gu, Yu and Zheng, Boyuan and Chen, Shijie and Stevens, Samuel and Wang, Boshi and Sun, Huan and Su, Yu},
  booktitle = {Advances in Neural Information Processing Systems (NeurIPS)},
  year      = {2023},
  note      = {arXiv:2306.06070}
}

@inproceedings{koh2024visualwebarena,
  title     = {{VisualWebArena: Evaluating Multimodal Agents on Realistic Visual Web Tasks}},
  author    = {Koh, Jing Yu and Lo, Robert and Jang, Lawrence and Duvvur, Vikram and Lim, Ming and Huang, Po-Yu and Neubig, Graham and Zhou, Shuyan and Salakhutdinov, Russ and Fried, Daniel},
  booktitle = {Proceedings of the 62nd Annual Meeting of the Association for Computational Linguistics (ACL)},
  year      = {2024},
  doi       = {10.18653/v1/2024.acl-long.50}
}

@inproceedings{lu2024weblinx,
  title     = {{WebLINX: Real-World Website Navigation with Multi-Turn Dialogue}},
  author    = {L\`u, Xing Han and Kasner, Zden\v{e}k and Reddy, Siva},
  booktitle = {International Conference on Machine Learning (ICML)},
  year      = {2024},
  note      = {arXiv:2402.05930}
}

@article{browsergym2025,
  title   = {{The BrowserGym Ecosystem for Web Agent Research}},
  author  = {Le Sellier De Chezelles, Thibault and Gasse, Maxime and Drouin, Alexandre and Caccia, Massimo and Boisvert, L\'eo and Thakkar, Megh and Marty, Tom and Assouel, Rim and Omidi Shayegan, Sahar and Jang, Lawrence Keunho and L\`u, Xing Han and Yoran, Ori and Kong, Dehan and Xu, Frank F. and Reddy, Siva and Cappart, Quentin and Neubig, Graham and Salakhutdinov, Ruslan and Chapados, Nicolas and Lacoste, Alexandre},
  journal = {Transactions on Machine Learning Research (TMLR)},
  year    = {2025},
  note    = {arXiv:2412.05467}
}

@article{pan2024webcanvas,
  title   = {{WebCanvas: Benchmarking Web Agents in Online Environments}},
  author  = {Pan, Yichen and Kong, Dehan and Zhou, Sida and Cui, Cheng and Leng, Yifei and Jiang, Bing and Liu, Hangyu and Shang, Yanyi and Zhou, Shuyan and Wu, Tongshuang and Wu, Zhengyang},
  journal = {arXiv preprint arXiv:2406.12373},
  year    = {2024}
}

@inproceedings{gou2025mind2web2,
  title     = {{Mind2Web 2: Evaluating Agentic Search with Agent-as-a-Judge}},
  author    = {Gou, Boyu and Huang, Zanming and Ning, Yuting and Gu, Yu and Lin, Michael and Qi, Weijian and Kopanev, Andrei and Yu, Botao and Jim\'enez Guti\'errez, Bernal and Shu, Yiheng and Song, Chan Hee and Wu, Jiaman and Chen, Shijie and Moussa, Hanane Nour and Zhang, Tianshu and Xie, Jian and Li, Yifei and Xue, Tianci and Liao, Zeyi and Zhang, Kai and Zheng, Boyuan and Cai, Zhaowei and Rozgic, Viktor and Ziyadi, Morteza and Sun, Huan and Su, Yu},
  booktitle = {Advances in Neural Information Processing Systems (NeurIPS)},
  year      = {2025}
}

@inproceedings{liu2025econwebarena,
  title     = {{EconWebArena: Benchmarking Autonomous Agents on Economic Tasks in Realistic Web Environments}},
  author    = {Liu, Zefang and Quan, Yinzhu},
  booktitle = {NeurIPS Workshop on Bridging Language, Agent, and World Models for Reasoning and Planning (LAW)},
  year      = {2025},
  note      = {arXiv:2506.08136}
}

@article{skillgen_ma2026,
  title   = {{SkillGen: Verified Inference-Time Agent Skill Synthesis}},
  author  = {Ma, Yuchen and Huang, Yue and Bao, Han and Zhuang, Haomin and Shukla, Swadheen and Galley, Michel and Zhang, Xiangliang and Feuerriegel, Stefan},
  journal = {arXiv preprint arXiv:2605.10999},
  year    = {2026}
}

@inproceedings{econlogicqa2024,
  title     = {{EconLogicQA: A Question-Answering Benchmark for Evaluating Large Language Models in Economic Sequential Reasoning}},
  author    = {Quan, Yinzhu and Liu, Zefang},
  booktitle = {Findings of the Association for Computational Linguistics: EMNLP 2024},
  year      = {2024},
  note      = {arXiv:2405.07938}
}

@article{invagent2024,
  title   = {{InvAgent: A Large Language Model based Multi-Agent System for Inventory Management in Supply Chains}},
  author  = {Quan, Yinzhu and Liu, Zefang},
  journal = {arXiv preprint arXiv:2407.11384},
  year    = {2024}
}

@article{mars2026,
  title   = {{Leveraging large language models to enhance multi-agent risk assessment in supply chain networks}},
  author  = {Quan, Yinzhu and Liu, Zefang and Benaben, Frederick and Montreuil, Benoit},
  journal = {International Journal of Production Research},
  year    = {2026},
  doi     = {10.1080/00207543.2026.2619562}
}

@inproceedings{crmagent2025,
  title     = {{CRMAgent: A Multi-Agent LLM System for E-Commerce CRM Message Template Generation}},
  author    = {Quan, Yinzhu and Li, Xinrui and Chen, Ying},
  booktitle = {GenAIECommerce: Workshop on Agentic and Generative AI for E-Commerce at RecSys},
  year      = {2025},
  note      = {arXiv:2507.08325}
}

@inproceedings{li2024econagent,
  title     = {{EconAgent: Large Language Model-Empowered Agents for Simulating Macroeconomic Activities}},
  author    = {Li, Nian and Gao, Chen and Li, Mingyu and Li, Yong and Liao, Qingmin},
  booktitle = {Proceedings of the 62nd Annual Meeting of the Association for Computational Linguistics},
  year      = {2024},
  doi       = {10.18653/v1/2024.acl-long.829}
}

@article{yang2024finrobot,
  title   = {{FinRobot: An Open-Source AI Agent Platform for Financial Applications using Large Language Models}},
  author  = {Yang, Hongyang and Zhang, Boyu and Wang, Neng and Guo, Cheng and Zhang, Xiaoli and Lin, Likun and Wang, Junlin and Zhou, Tianyu and Guan, Mao and Zhang, Runjia and Wang, Christina Dan},
  journal = {arXiv preprint arXiv:2405.14767},
  year    = {2024}
}

@article{voyager2023,
  title   = {{Voyager: An Open-Ended Embodied Agent with Large Language Models}},
  author  = {Wang, Guanzhi and Xie, Yuqi and Jiang, Yunfan and Mandlekar, Ajay and Xiao, Chaowei and Zhu, Yuke and Fan, Linxi and Anandkumar, Anima},
  journal = {Transactions on Machine Learning Research (TMLR)},
  year    = {2024},
  note    = {arXiv:2305.16291}
}

@inproceedings{webvoyager2024,
  title     = {{WebVoyager: Building an End-to-End Web Agent with Large Multimodal Models}},
  author    = {He, Hongliang and Yao, Wenlin and Ma, Kaixin and Yu, Wenhao and Dai, Yong and Zhang, Hongming and Lan, Zhenzhong and Yu, Dong},
  booktitle = {Proceedings of the 62nd Annual Meeting of the Association for Computational Linguistics (ACL)},
  year      = {2024},
  note      = {arXiv:2401.13919}
}

@inproceedings{seeact2024,
  title     = {{GPT-4V(ision) is a Generalist Web Agent, if Grounded}},
  author    = {Zheng, Boyuan and Gou, Boyu and Kil, Jihyung and Sun, Huan and Su, Yu},
  booktitle = {International Conference on Machine Learning (ICML)},
  year      = {2024},
  note      = {arXiv:2401.01614}
}

@inproceedings{reflexion2023,
  title     = {{Reflexion: Language Agents with Verbal Reinforcement Learning}},
  author    = {Shinn, Noah and Cassano, Federico and Berman, Edward and Gopinath, Ashwin and Narasimhan, Karthik and Yao, Shunyu},
  booktitle = {Advances in Neural Information Processing Systems (NeurIPS)},
  year      = {2023},
  note      = {arXiv:2303.11366}
}

@inproceedings{zhao2024expel,
  title     = {{ExpeL: LLM Agents Are Experiential Learners}},
  author    = {Zhao, Andrew and Huang, Daniel and Xu, Quentin and Lin, Matthieu and Liu, Yong-Jin and Huang, Gao},
  booktitle = {Proceedings of the AAAI Conference on Artificial Intelligence},
  year      = {2024},
  volume    = {38},
  doi       = {10.1609/aaai.v38i17.29936}
}

@inproceedings{wang2025awm,
  title     = {{Agent Workflow Memory}},
  author    = {Wang, Zora Zhiruo and Mao, Jiayuan and Fried, Daniel and Neubig, Graham},
  booktitle = {Proceedings of the 42nd International Conference on Machine Learning},
  year      = {2025},
  volume    = {267},
  series    = {Proceedings of Machine Learning Research}
}

@inproceedings{kim2024rada,
  title     = {{RaDA: Retrieval-Augmented Web Agent Planning with LLMs}},
  author    = {Kim, Minsoo and Bursztyn, Victor and Koh, Eunyee and Guo, Shunan and Hwang, Seung-won},
  booktitle = {Findings of the Association for Computational Linguistics: ACL 2024},
  year      = {2024},
  doi       = {10.18653/v1/2024.findings-acl.802}
}

@article{jang2024videowebarena,
  title   = {{VideoWebArena: Evaluating Long Context Multimodal Agents with Video Understanding Web Tasks}},
  author  = {Jang, Lawrence and Li, Yinheng and Zhao, Dan and Ding, Charles and Lin, Justin and Liang, Paul Pu and Bonatti, Rogerio and Koishida, Kazuhito},
  journal = {arXiv preprint arXiv:2410.19100},
  year    = {2024}
}

@article{zheng2025skillweaver,
  title   = {{SkillWeaver: Web Agents Can Self-Improve by Discovering and Honing Skills}},
  author  = {Zheng, Boyuan and Fatemi, Michael Y. and Jin, Xiaolong and Wang, Zora Zhiruo and Gandhi, Apurva and Song, Yueqi and Gu, Yu and Srinivasa, Jayanth and Liu, Gaowen and Neubig, Graham and Su, Yu},
  journal = {arXiv preprint arXiv:2504.07079},
  year    = {2025}
}

@article{contractskill2026,
  title   = {{ContractSkill: Repairable Contract-Based Skills for Multimodal Web Agents}},
  author  = {Lu, Zijian and Zuo, Yiping and Nie, Yupeng and He, Xin and Fan, Weibei and Dai, Chen},
  journal = {arXiv preprint arXiv:2603.20340},
  year    = {2026}
}

@article{polyskill2025,
  title   = {{PolySkill: Learning Generalizable Skills Through Polymorphic Abstraction}},
  author  = {Yu, Simon and Li, Gang and Shi, Weiyan and Qi, Peng},
  journal = {arXiv preprint arXiv:2510.15863},
  year    = {2025}
}

@article{skillsbench2026,
  title   = {{SkillsBench: Benchmarking How Well Agent Skills Work Across Diverse Tasks}},
  author  = {Li, Xiangyi and Chen, Wenbo and Liu, Yimin and Zheng, Shenghan and Chen, Xiaokun and He, Yifeng and others},
  journal = {arXiv preprint arXiv:2602.12670},
  year    = {2026}
}

@article{mcnemar1947,
  title   = {{Note on the Sampling Error of the Difference Between Correlated Proportions or Percentages}},
  author  = {McNemar, Quinn},
  journal = {Psychometrika},
  year    = {1947},
  volume  = {12},
  number  = {2},
  doi     = {10.1007/BF02295996}
}

\normalsize
\appendix
\raggedbottom
\makeatletter
\setlength{\@fptop}{0pt}
\setlength{\@fpsep}{12pt plus 2pt}
\setlength{\@fpbot}{0pt plus 1fil}
\makeatother
\section{Reproduction Details}
\label{app:repro}

The \nSource{} EconSkills procedures are available on
Hugging Face\footnote{\url{https://huggingface.co/datasets/EconWebArena/EconSkills}}. Each JSON file contains the seven fields
defined in Section~\ref{sec:method}: name, purpose, preconditions, procedure,
website guidance, verification, and recovery. The same fields are rendered to
Markdown when a skill is placed in the agent prompt. Appendix~\ref{app:skill-example}
shows two examples from the library in full.

The experiments can be rerun on EconWebArena through BrowserGym and AgentLab
using the agent configuration in Section~\ref{sec:setup} and the prompt
conditions in Section~\ref{sec:application}. The controlled study evaluates
\nVariant{} variants under 3 conditions and 3 runs per condition. The deployment
study evaluates each condition once on all \nTasks{} tasks. Both studies use the
native benchmark reward: the returned answer must contain the gold value and the
browser must finish on the authoritative domain.

Four benchmark values that had changed on their authoritative websites are
handled through a fixed task-level correction table constructed without reference
to condition outcomes. The HINT30 prompt states that the supplied skill may be
incomplete, outdated, or mismatched and should be ignored when it does not fit
the visible page. Its 150 direct-match tasks receive the corresponding family
skill, while the remaining 210 tasks receive the top-1 retrieved skill.

\section{Skill Examples}
\label{app:skill-example}

The first example is the ECB skill summarized in Table~\ref{tab:example-skill};
the second covers a different workflow on the USCIS fee schedule. Both preserve
the seven fields and operational content of the stored JSON skills. Angle-bracketed
terms denote values supplied by a new task rather than copied from the source
trajectory.

\subsection{ECB Reference Rate}

\begin{framed}
\raggedright
\textbf{Name.} \texttt{lookup\_ecb\_euro\_reference\_fx\_rate}

\textbf{Purpose.} Retrieve the ECB euro foreign-exchange reference rate for
a specified currency and date.

\textbf{Preconditions.} The task requests a euro-to-currency exchange rate
from the European Central Bank; a target calendar date is provided or
inferable; and the ECB website is reachable.

\textbf{Procedure.}
\begin{enumerate}
  \item Open the ECB ``Euro foreign exchange reference rates'' section.
  \item Navigate to the target-currency page. ECB quotes units of the target
  currency per euro.
  \item Set the start and end of the date range to \texttt{<date>}.
  \item Switch to a table or another exact-value view instead of relying on a
  chart tooltip.
  \item Read the rate for \texttt{<date>} and preserve four decimal places.
\end{enumerate}

\textbf{Website guidance.} Currency pages follow the pattern
\texttt{eurofxref-graph-<currency>.en.html}. The date inputs use
day/month/year; setting both fields to the same day isolates one observation.
Use ``Show table'' or the CSV download when the chart is ambiguous.

\textbf{Verification.} Confirm the ECB domain and page heading, the requested
currency, and the target-currency-per-EUR orientation. Check that the displayed
date exactly matches \texttt{<date>} and that the value is a nominal reference
rate rather than a percentage change or index.

\textbf{Recovery.} If the currency page is missing, return to the landing
page or use ECB site search. If the chart does not expose an exact value, use the
table or CSV view. If date entry fails, use the date picker. For a non-business
day, do not silently substitute the previous observation unless the task permits
it. If the layout changes, navigate from the ECB home page through
Stats $\rightarrow$ Policy and exchange rates, or use site search.
\end{framed}

\subsection{USCIS Fee Schedule}

\begin{framed}
\raggedright
\textbf{Name.} \texttt{lookup\_uscis\_fee\_schedule}

\textbf{Purpose.} Retrieve a USCIS fee for a specified form or process and
effective date from the official fee schedule on \texttt{uscis.gov}.

\textbf{Preconditions.} The task identifies a form or process, any relevant
fee qualifier such as per beneficiary or per filing, and an effective date. The
answer must come from an authoritative USCIS page and use the requested dollar
format.

\textbf{Procedure.}
\begin{enumerate}
  \item Open the USCIS Fee Schedule and confirm that its effective date matches
  \texttt{<date>}.
  \item Use the schedule filters, search field, or find-in-page to locate
  \texttt{<form-or-process>}.
  \item Read the full matching entry and select the line corresponding to
  \texttt{<fee-qualifier>} and the requested operation.
  \item Read the listed fee and check any notes, surcharges, or exemptions that
  could change its interpretation.
  \item Normalize the amount to the output format requested by the task.
\end{enumerate}

\textbf{Website guidance.} USCIS maintains the schedule at
\texttt{uscis.gov/g-1055}. Its effective date appears near the top of the page,
and topic filters or a search box may appear above the table. If a filter does
not respond, clear it and use find-in-page with the form number or process name.
News and rulemaking pages may explain a fee change, but the requested amount
should be read from the applicable fee schedule.

\textbf{Verification.} Confirm the \texttt{uscis.gov} domain, the fee
schedule title, and the effective date. Check that the selected row matches the
form, operation, and qualifier exactly, and that the value is the requested fee
rather than a separate surcharge, discounted amount, or exemption. When several
rows have similar names, compare their full descriptions before selecting one.

\textbf{Recovery.} If the schedule does not load, use USCIS site search for
``Fee Schedule'' or ``G-1055.'' If the displayed schedule has the wrong effective
date, follow its links to the applicable version. If dynamic filters fail, reload
the page, use a text or print view when available, or search the unfiltered table.
When official USCIS pages show conflicting amounts, prioritize the fee schedule
whose effective date matches the task.
\end{framed}

\section{Statistical Aggregation Details}
\label{app:audit}

The controlled result table is indexed by source task, held-out variant,
condition, and seed. This gives $50\times2\times3\times3=900$ cells. For each
cell, we use the native benchmark reward. The 300 matched BASE/MATCH1 cells yield 46 failure-to-success changes and
21 success-to-failure changes. At the source-variant level, we average the three
runs within each condition before assigning the four outcome patterns in
Table~\ref{tab:pair-outcomes}; this prevents one unusually easy seed from being
reported as a separate task family.

For the full-library deployment study, each condition has one row for each of the 360 tasks.
The HINT30 direct-match label is determined by task construction: the 50 original
source tasks and 100 variants have a direct family skill, whereas the other 210
tasks do not. The latter still receive a retrieved hint, so ``no direct match''
does not mean an empty prompt. Pairwise tests use the BASE and treatment reward
for the same task. Because there is only one run per task, we do not report
across-seed error bars for this study.

\section{Qualitative Browser Cases}
\label{app:cases}

\begin{figure}[!htbp]
  \centering
  \setlength\fboxsep{1pt}
  \setlength\fboxrule{0.5pt}
  \begin{subfigure}[t]{0.32\linewidth}
    \centering\fbox{\includegraphics[width=.98\linewidth]{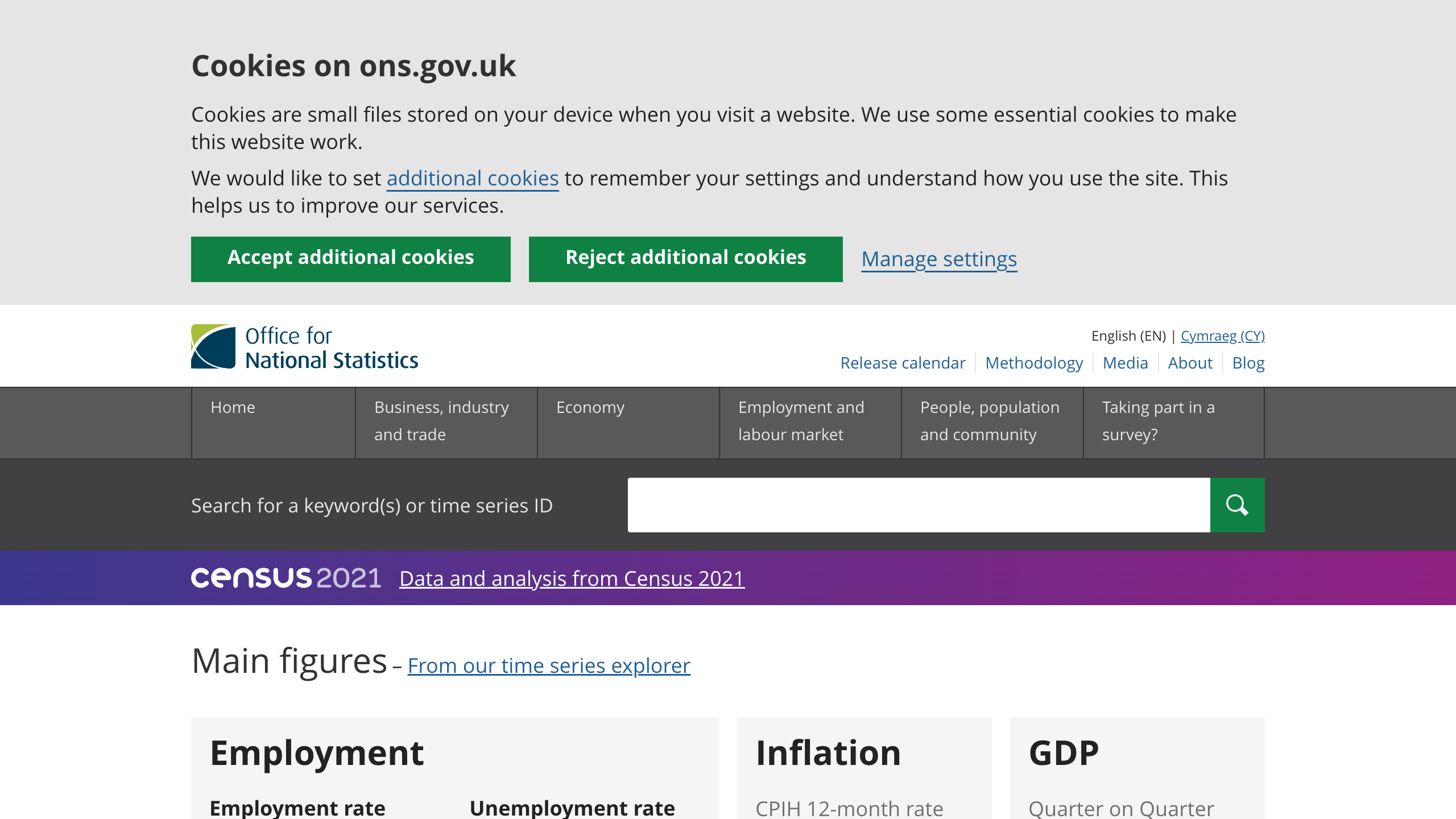}}
    \caption{BASE, step 0: ONS home}
  \end{subfigure}\hfill
  \begin{subfigure}[t]{0.32\linewidth}
    \centering\fbox{\includegraphics[width=.98\linewidth]{figures/ons_base_step15.png}}
    \caption{BASE, step 15: dataset page}
  \end{subfigure}\hfill
  \begin{subfigure}[t]{0.32\linewidth}
    \centering\fbox{\includegraphics[width=.98\linewidth]{figures/ons_base_step30.png}}
    \caption{BASE, step 30: download view}
  \end{subfigure}

  \vspace{5pt}
  \begin{subfigure}[t]{0.32\linewidth}
    \centering\fbox{\includegraphics[width=.98\linewidth]{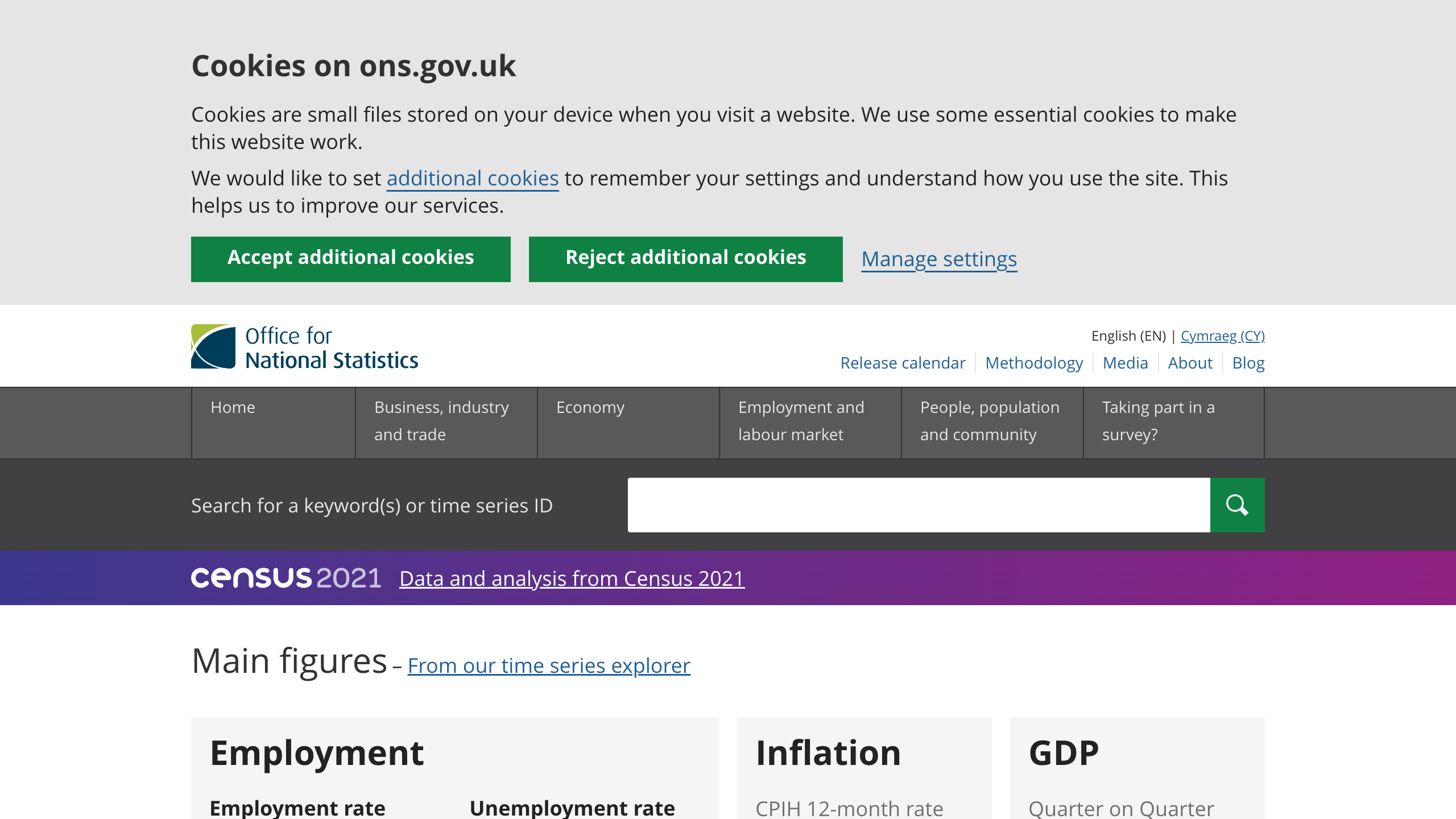}}
    \caption{HINT30, step 0: ONS home}
  \end{subfigure}\hfill
  \begin{subfigure}[t]{0.32\linewidth}
    \centering\fbox{\includegraphics[width=.98\linewidth]{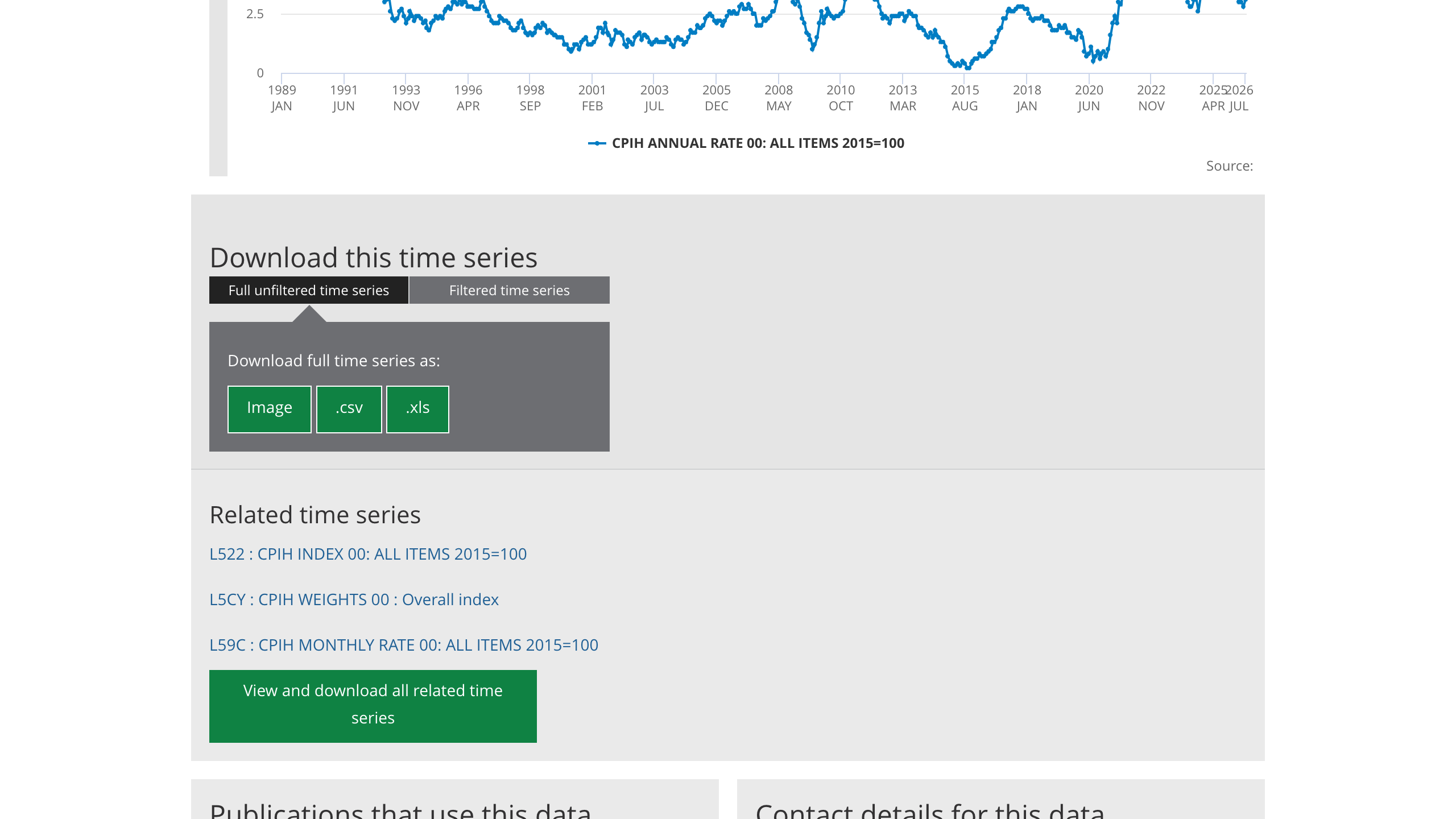}}
    \caption{HINT30, step 7: series page}
  \end{subfigure}\hfill
  \begin{subfigure}[t]{0.32\linewidth}
    \centering\fbox{\includegraphics[width=.98\linewidth]{figures/ons_hint_step14.png}}
    \caption{HINT30, step 14: answer table}
  \end{subfigure}
  \caption{Browser trajectories for Task 225 on ONS. BASE remains at a general
  dataset download interface after 30 steps. With the matched optional hint, the agent reaches the
  dedicated MM23 time-series table and succeeds in 14 steps.}
  \label{fig:ons-case}
\end{figure}

\begin{figure}[!htbp]
  \centering
  \setlength\fboxsep{1pt}
  \setlength\fboxrule{0.5pt}
  \begin{subfigure}[t]{0.32\linewidth}
    \centering\fbox{\includegraphics[width=.98\linewidth]{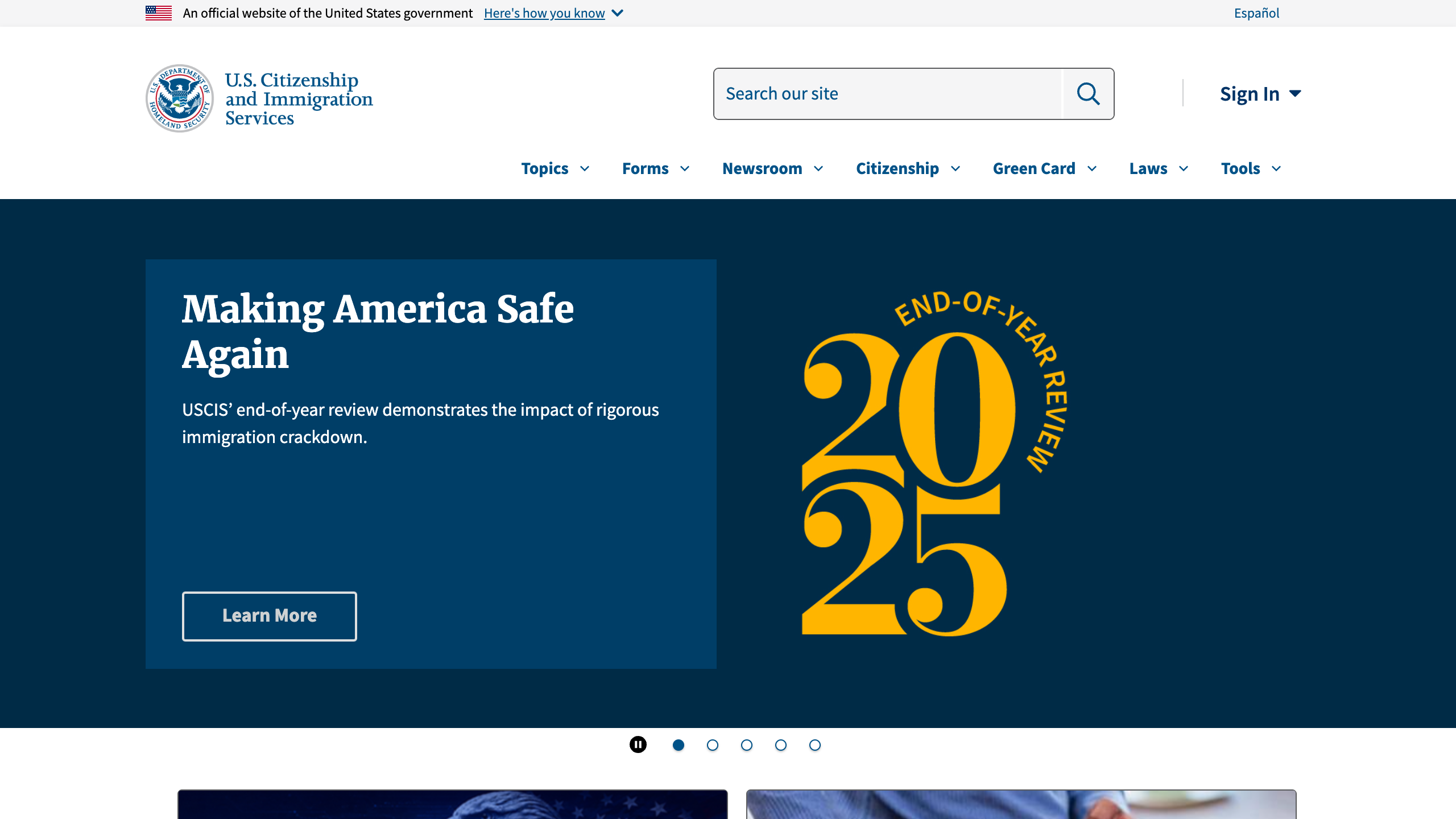}}
    \caption{BASE, step 0: USCIS home}
  \end{subfigure}\hfill
  \begin{subfigure}[t]{0.32\linewidth}
    \centering\fbox{\includegraphics[width=.98\linewidth]{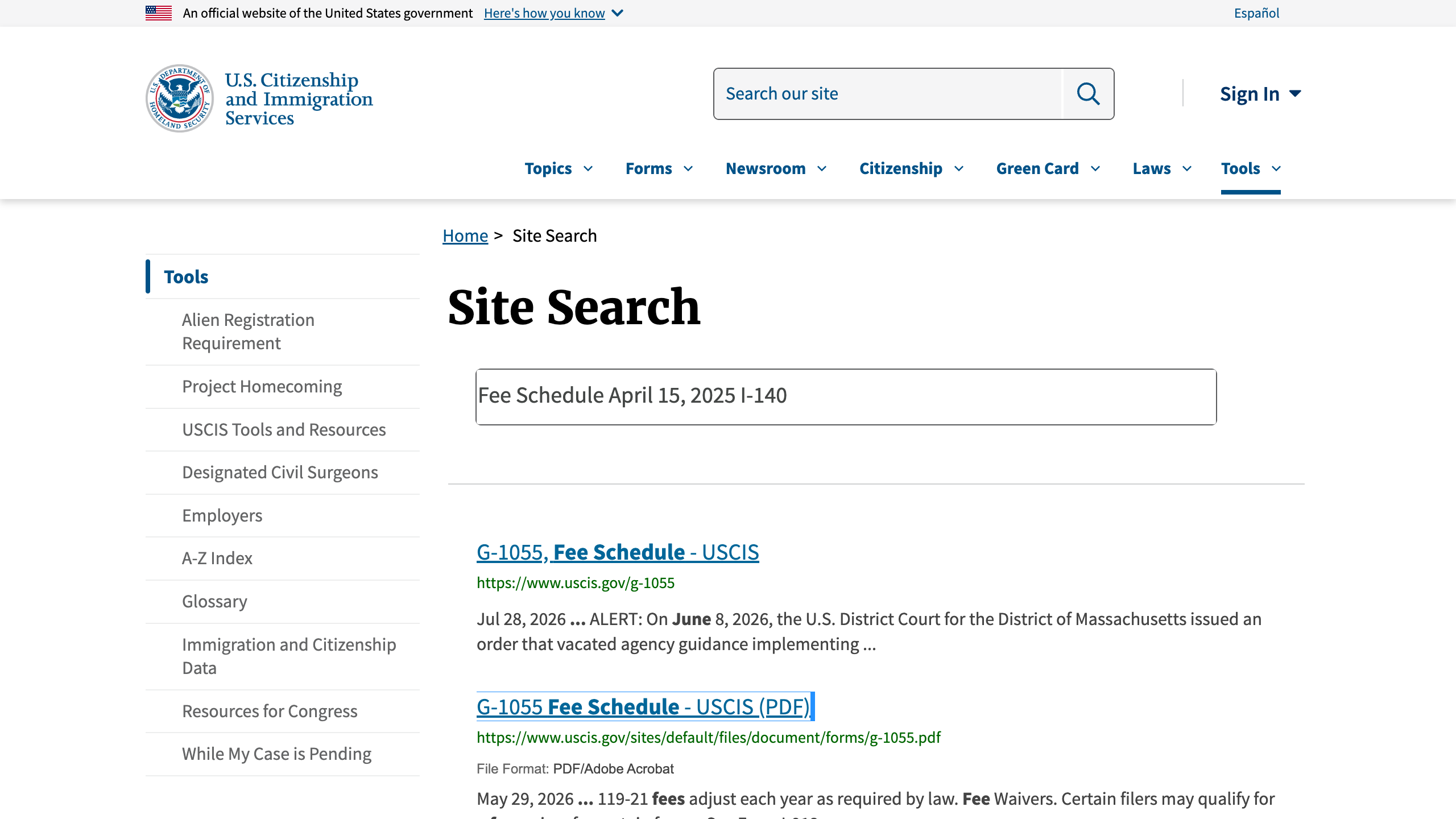}}
    \caption{BASE, step 5: fee search}
  \end{subfigure}\hfill
  \begin{subfigure}[t]{0.32\linewidth}
    \centering\fbox{\includegraphics[width=.98\linewidth]{figures/uscis_base_step9.png}}
    \caption{BASE, step 9: answer table}
  \end{subfigure}

  \vspace{5pt}
  \begin{subfigure}[t]{0.32\linewidth}
    \centering\fbox{\includegraphics[width=.98\linewidth]{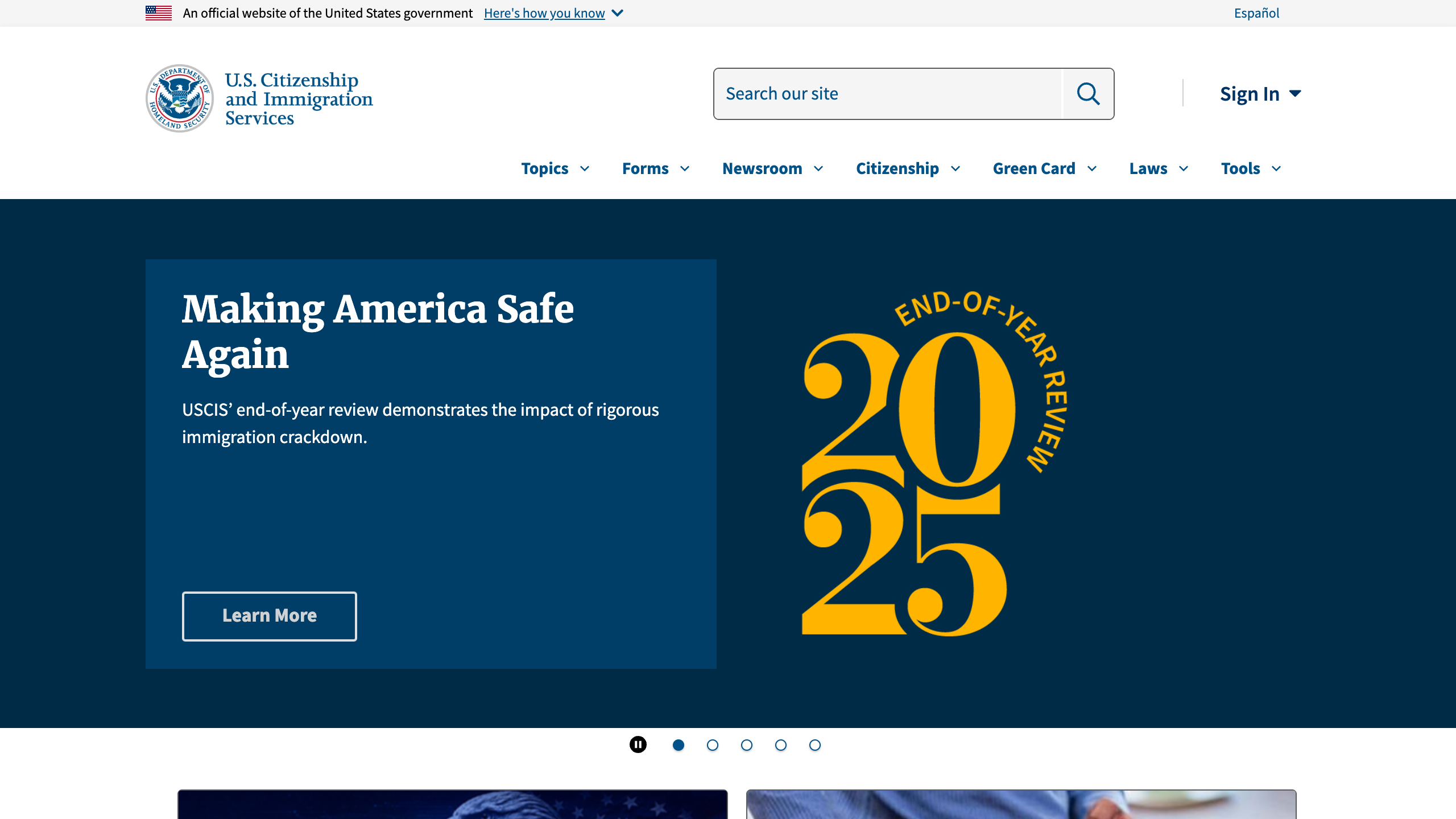}}
    \caption{HINT30, step 0: USCIS home}
  \end{subfigure}\hfill
  \begin{subfigure}[t]{0.32\linewidth}
    \centering\fbox{\includegraphics[width=.98\linewidth]{figures/uscis_hint_step15.png}}
    \caption{HINT30, step 15: fee search}
  \end{subfigure}\hfill
  \begin{subfigure}[t]{0.32\linewidth}
    \centering\fbox{\includegraphics[width=.98\linewidth]{figures/uscis_hint_step30.png}}
    \caption{HINT30, step 30: I-140 section}
  \end{subfigure}
  \caption{Browser trajectories for Task 264 on USCIS. BASE reaches the I-140 fee table and succeeds in
  9 steps. HINT30 reaches the relevant section but does not return the verified
  value within 30 steps, separating correct navigation from correct extraction.}
  \label{fig:uscis-case}
\end{figure}

\begin{figure}[!htbp]
  \centering
  \setlength\fboxsep{1pt}
  \setlength\fboxrule{0.5pt}
  \begin{subfigure}[t]{0.32\linewidth}
    \centering\fbox{\includegraphics[width=.98\linewidth]{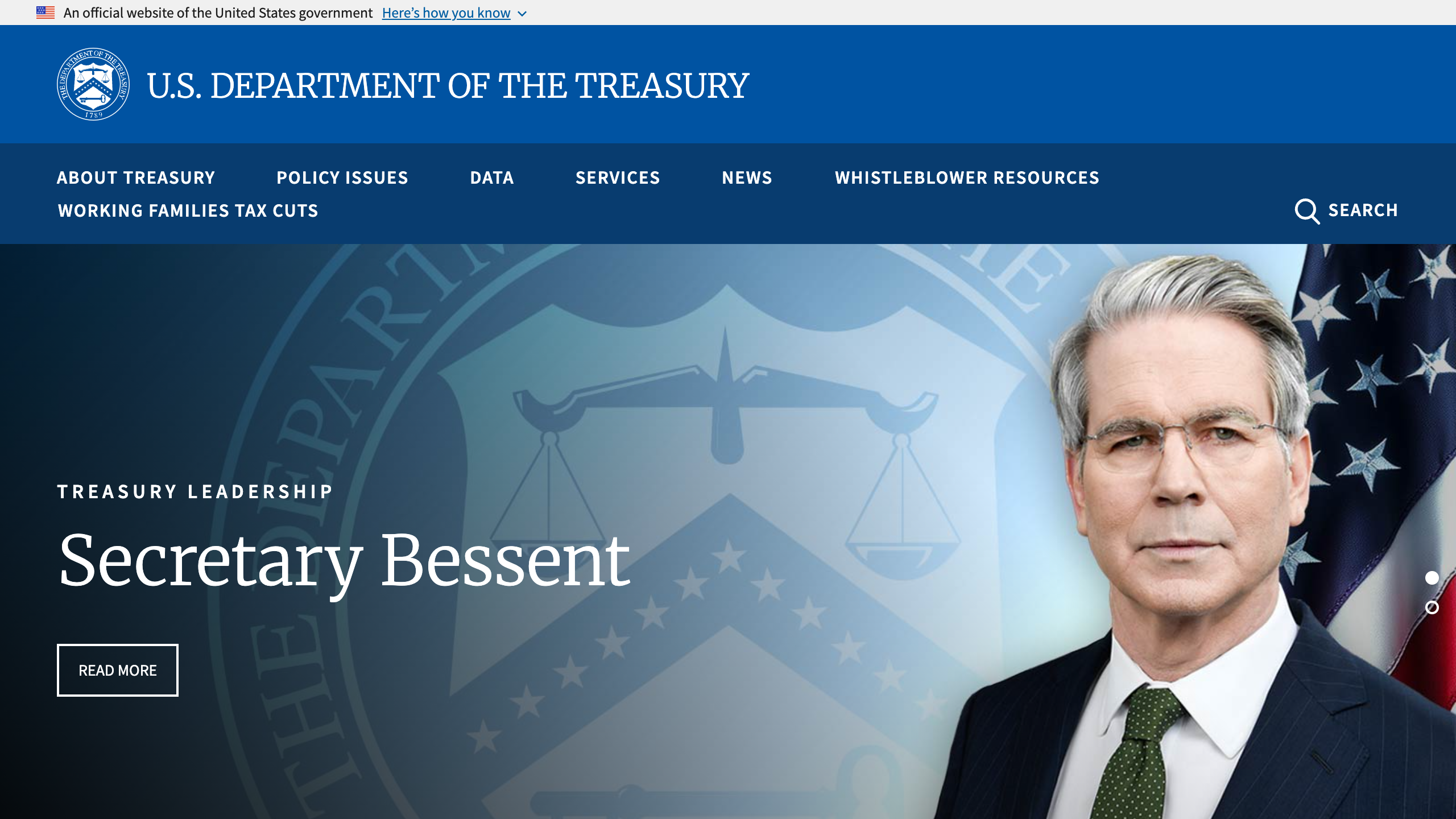}}
    \caption{BASE, step 0: Treasury home}
  \end{subfigure}\hfill
  \begin{subfigure}[t]{0.32\linewidth}
    \centering\fbox{\includegraphics[width=.98\linewidth]{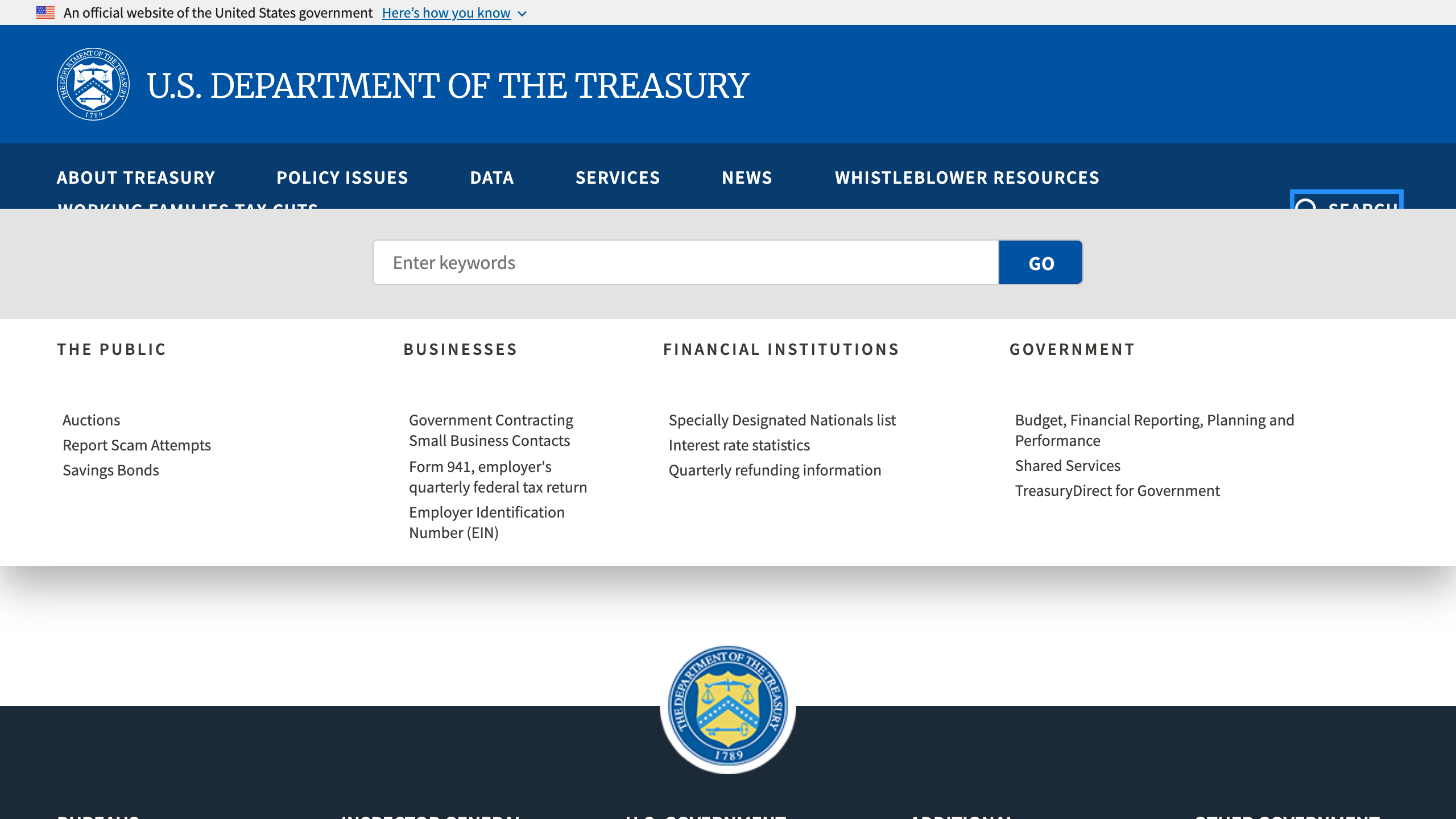}}
    \caption{BASE, step 15: site search}
  \end{subfigure}\hfill
  \begin{subfigure}[t]{0.32\linewidth}
    \centering\fbox{\includegraphics[width=.98\linewidth]{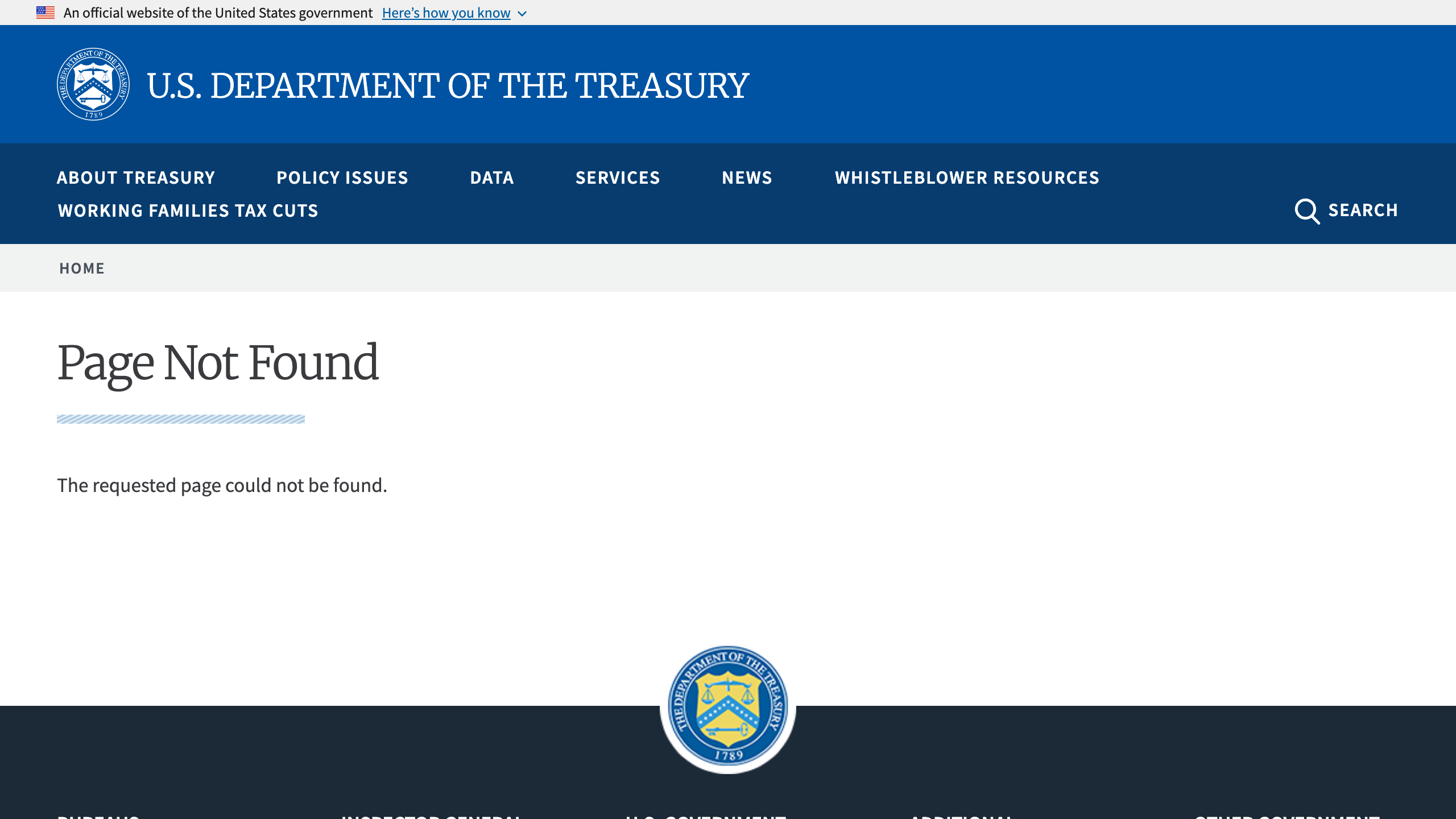}}
    \caption{BASE, step 30: dead link}
  \end{subfigure}

  \vspace{5pt}
  \begin{subfigure}[t]{0.32\linewidth}
    \centering\fbox{\includegraphics[width=.98\linewidth]{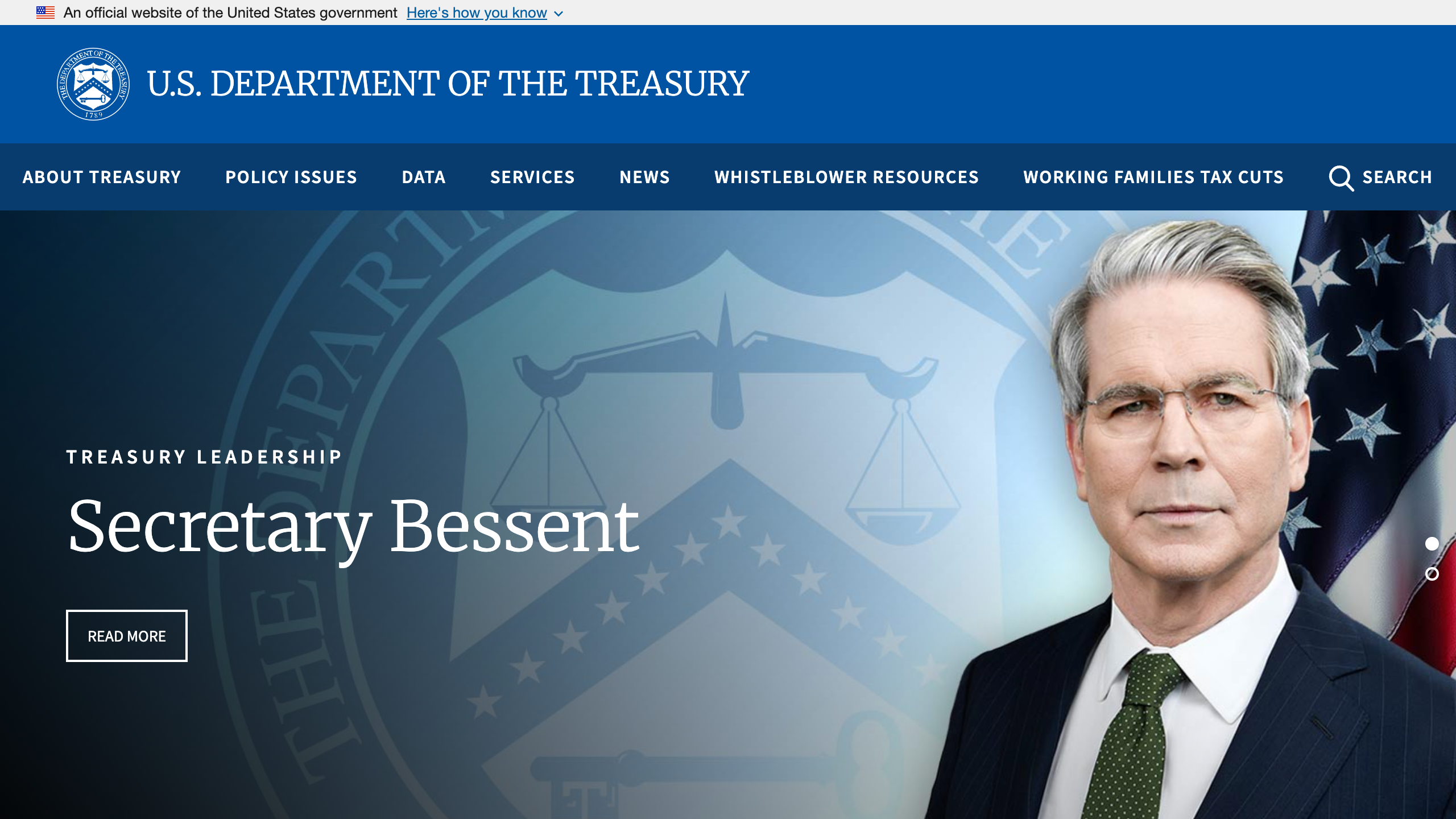}}
    \caption{HINT30, step 0: Treasury home}
  \end{subfigure}\hfill
  \begin{subfigure}[t]{0.32\linewidth}
    \centering\fbox{\includegraphics[width=.98\linewidth]{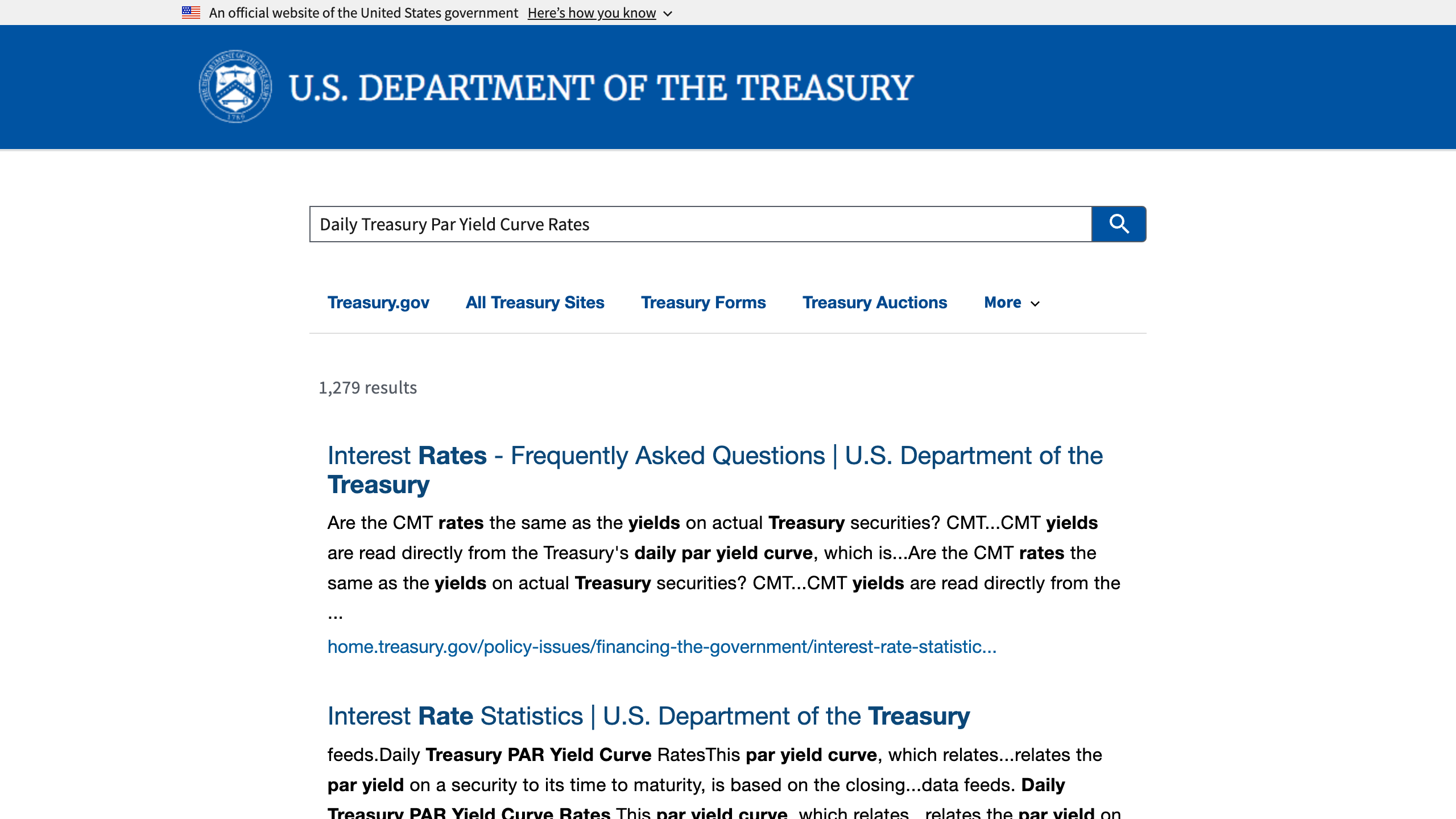}}
    \caption{HINT30, step 8: targeted search}
  \end{subfigure}\hfill
  \begin{subfigure}[t]{0.32\linewidth}
    \centering\fbox{\includegraphics[width=.98\linewidth]{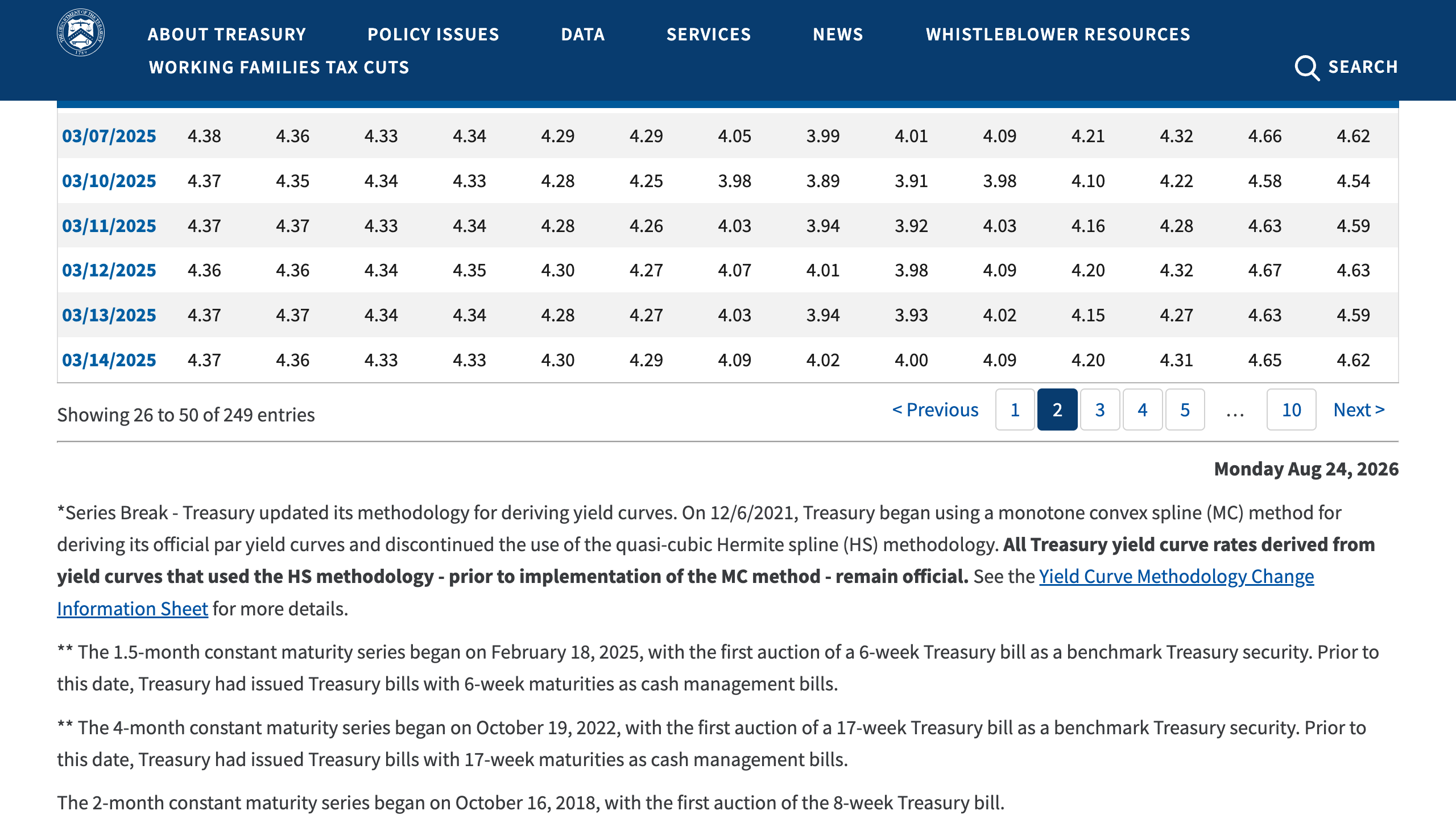}}
    \caption{HINT30, step 15: yield table}
  \end{subfigure}
  \caption{Browser trajectories for Task 274 on the U.S. Treasury site. BASE remains in site search and
  ends on a dead link after 30 steps. The matched hint names the daily par-yield
  route and its recovery strategy; the agent reaches the historical table and
  succeeds in 15 steps.}
  \label{fig:treasury-case}
\end{figure}

\begin{figure}[!htbp]
  \centering
  \setlength\fboxsep{1pt}
  \setlength\fboxrule{0.5pt}
  \begin{subfigure}[t]{0.32\linewidth}
    \centering\fbox{\includegraphics[width=.98\linewidth]{figures/ons_base_step30.png}}
    \caption{ONS BASE, step 30: unresolved}
  \end{subfigure}\hfill
  \begin{subfigure}[t]{0.32\linewidth}
    \centering\fbox{\includegraphics[width=.98\linewidth]{figures/uscis_base_step9.png}}
    \caption{USCIS BASE, step 9: success}
  \end{subfigure}\hfill
  \begin{subfigure}[t]{0.32\linewidth}
    \centering\fbox{\includegraphics[width=.98\linewidth]{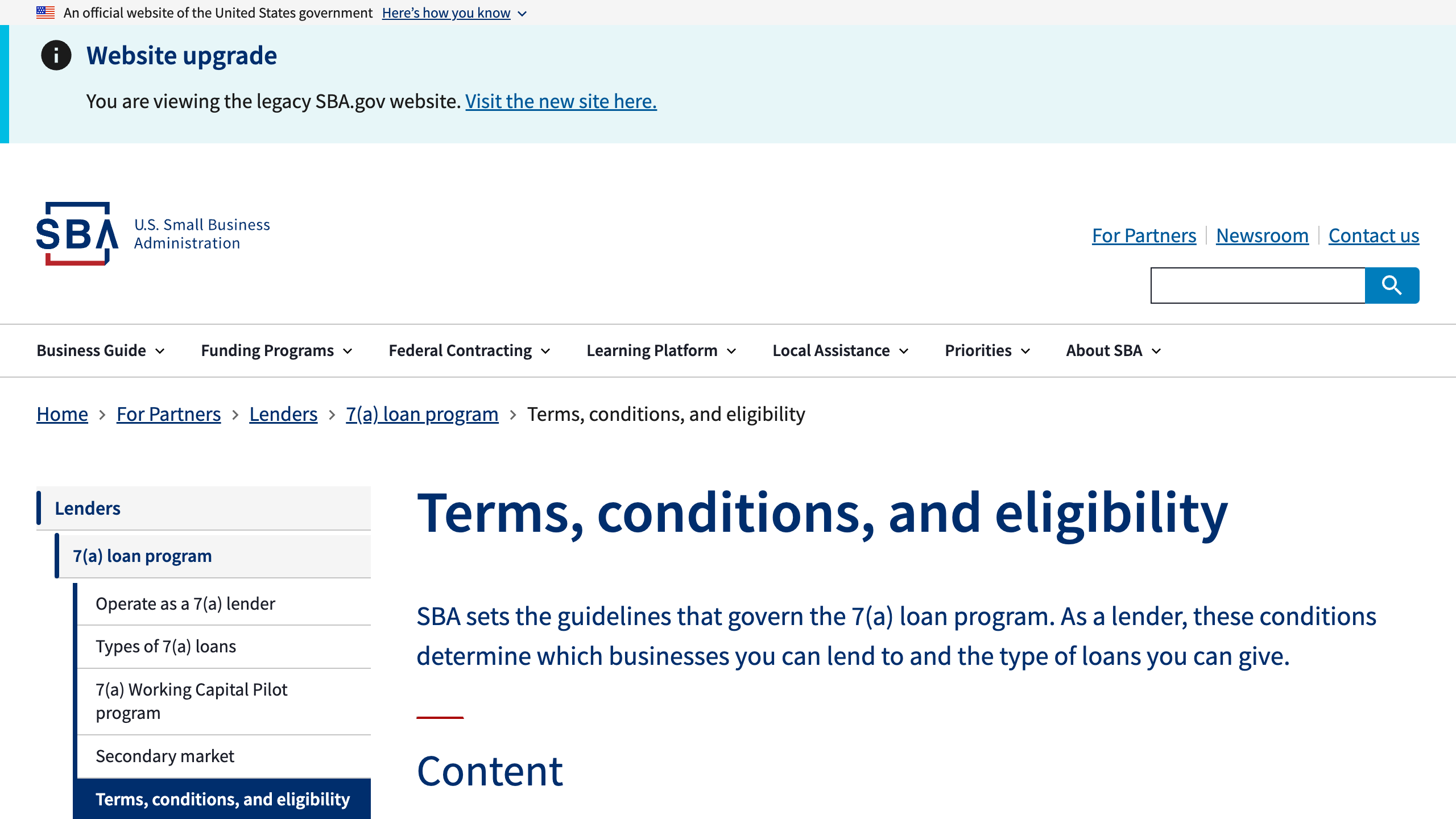}}
    \caption{SBA BASE, step 12: success}
  \end{subfigure}

  \vspace{5pt}
  \begin{subfigure}[t]{0.32\linewidth}
    \centering\fbox{\includegraphics[width=.98\linewidth]{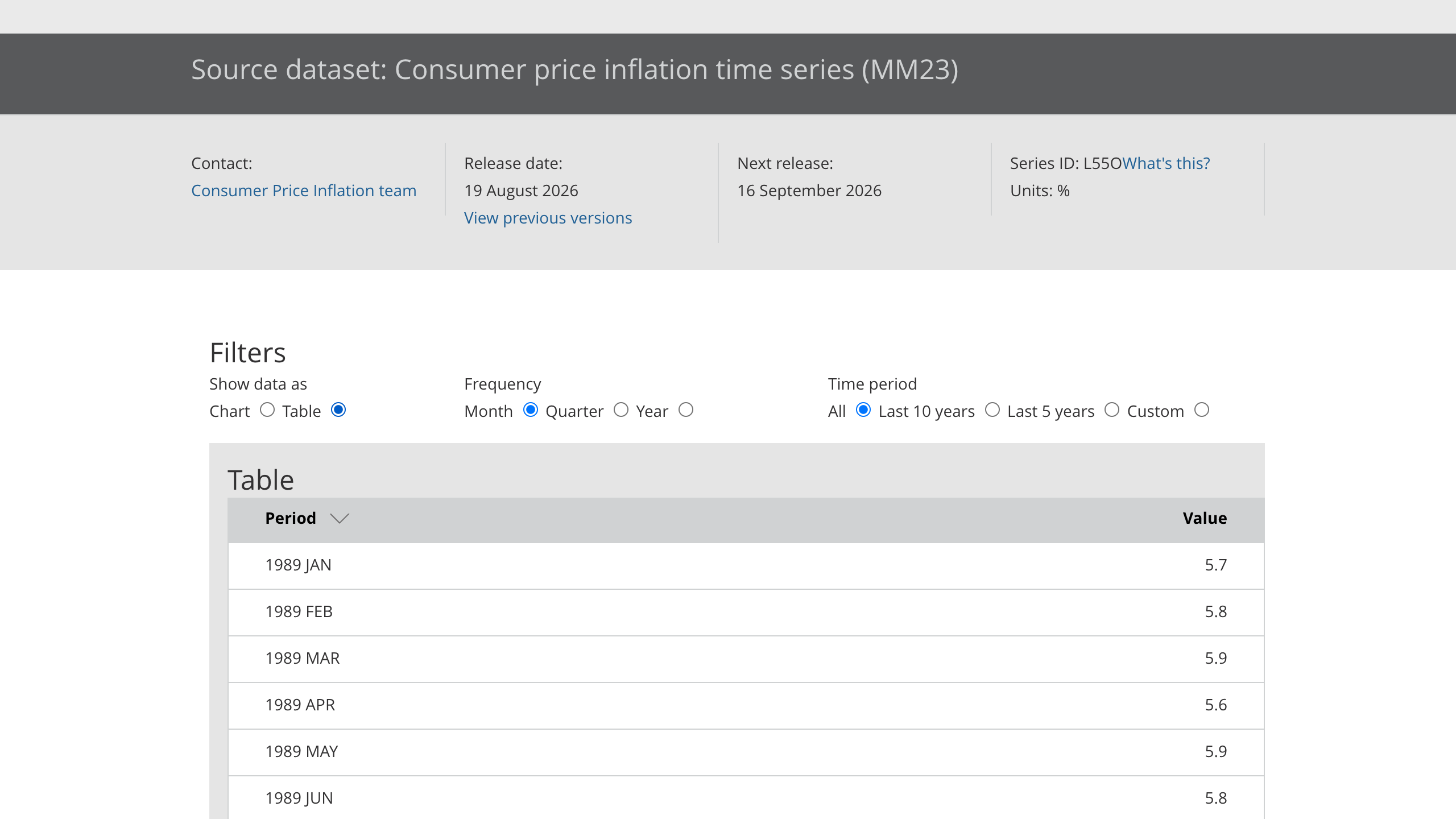}}
    \caption{ONS RETR5, step 5: success}
  \end{subfigure}\hfill
  \begin{subfigure}[t]{0.32\linewidth}
    \centering\fbox{\includegraphics[width=.98\linewidth]{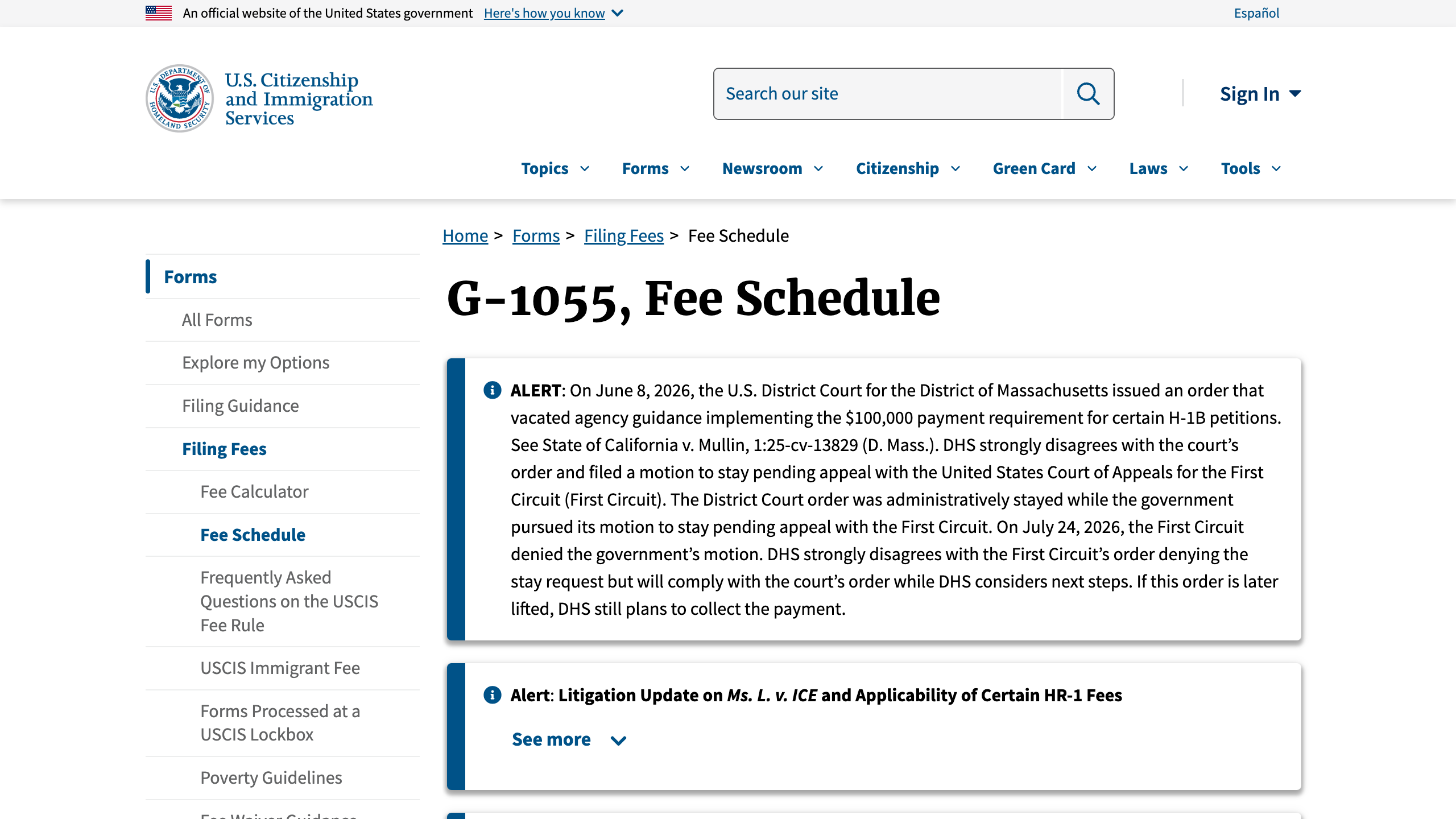}}
    \caption{USCIS RETR5, step 30: unresolved}
  \end{subfigure}\hfill
  \begin{subfigure}[t]{0.32\linewidth}
    \centering\fbox{\includegraphics[width=.98\linewidth]{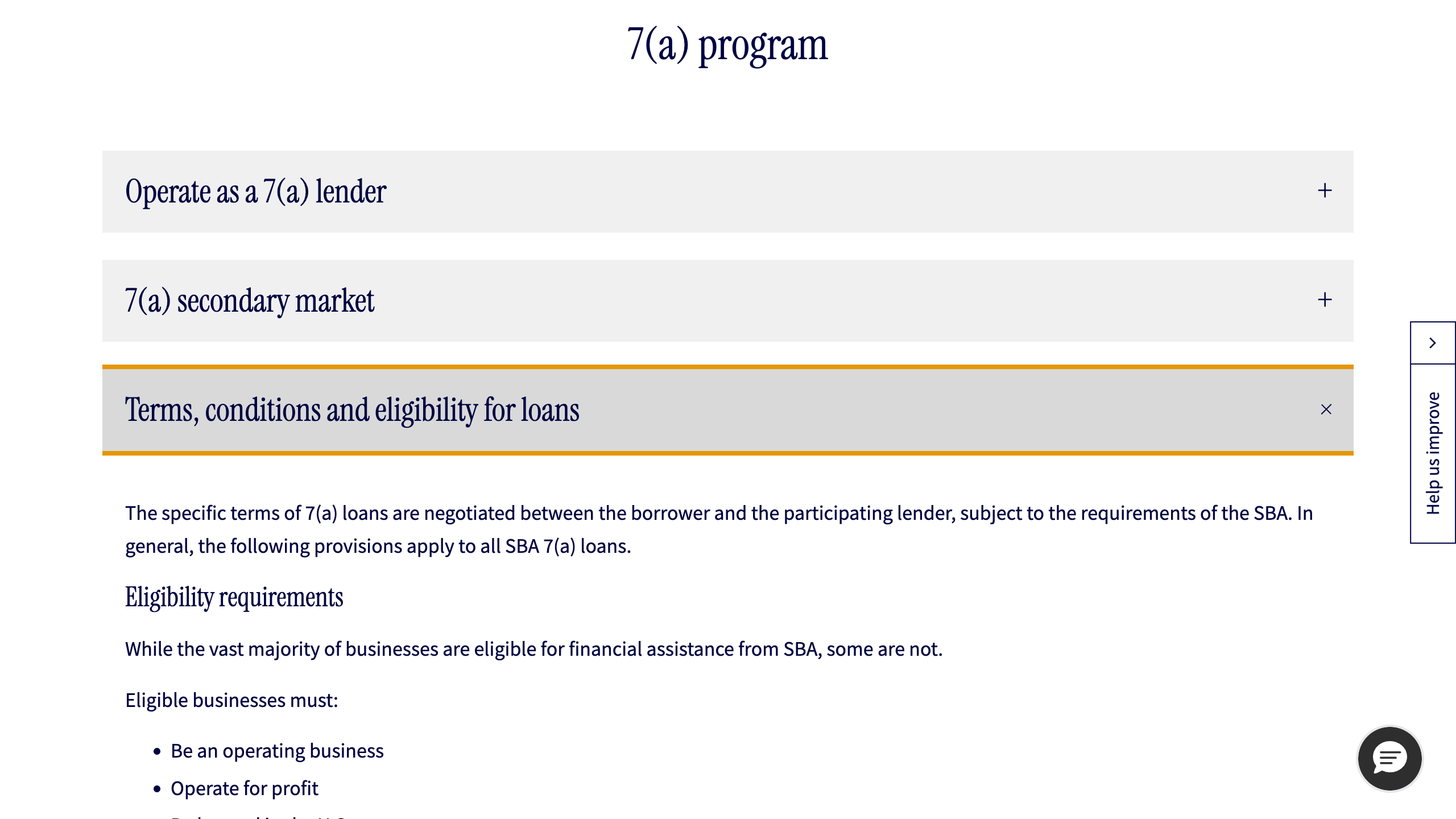}}
    \caption{SBA RETR5, step 3: success}
  \end{subfigure}
  \caption{BASE and RETR5 endpoints on three economic web tasks. RETR5 converts
  the ONS endpoint into a 5-step success and reaches the SBA evidence in 3 rather
  than 12 steps. The USCIS pair shows that reaching the correct fee page still
  requires exact extraction. Together, the cases illustrate retrieval gains and
  the role of semantic verification; aggregate performance is reported in
  Table~\ref{tab:library}.}
  \label{fig:retr5-cases}
\end{figure}

Figures~\ref{fig:ons-case} to~\ref{fig:treasury-case} compare browser paths from
three full-library deployment cases. The ONS and Treasury cases show the positive mechanism:
a suitable skill names a portal-specific destination and recovery route, which
avoids broad download interfaces or dead links
(Figures~\ref{fig:ons-case} and~\ref{fig:treasury-case}). The USCIS comparison
separates navigation from answer validation: both agents reach relevant content,
but the hinted run does not complete extraction and verification
(Figure~\ref{fig:uscis-case}). Figure~\ref{fig:retr5-cases} then compares BASE
and RETR5 endpoints on ONS, USCIS, and SBA tasks. RETR5 shortens the successful
ONS and SBA runs, while the USCIS pair isolates extraction as a separate issue.
Together, the cases show that skills can reduce route-finding effort and that
economic values still require explicit semantic checks.

\end{document}